\documentclass{article}

\usepackage{microtype}
\usepackage{graphicx}
\usepackage{subfigure}
\usepackage{booktabs} 

\usepackage{hyperref}

\usepackage[accepted]{icml2025}

\usepackage{amsmath}
\usepackage{amssymb}
\usepackage{mathtools}
\usepackage{amsthm}

\usepackage[capitalize,noabbrev]{cleveref}

\usepackage{url}
\usepackage{graphicx}
\usepackage{amsmath}
\usepackage{amssymb}
\usepackage[font=normalsize]{caption}
\usepackage{textcomp}
\usepackage{float}

\usepackage{subcaption}
\usepackage{xcolor}
\usepackage{booktabs}
\usepackage{makecell}
\usepackage{colortbl}
\definecolor{deepred}{rgb}{0, 0.3, 0.7}
\definecolor{Gray}{gray}{0.9}
\newcolumntype{g}{>{\columncolor{Gray}}c}
\definecolor{LightCyan}{rgb}{0.75,1,1}
\newcolumntype{y}{>{\columncolor{LightCyan}}c}
\definecolor{LightMagenta}{rgb}{1, 0.8, 0.8}
\newcolumntype{a}{>{\columncolor{LightMagenta}}c}

\usepackage[many]{tcolorbox}  
\newtcolorbox{boxL}{
    fontupper = \color{black},
    rounded corners,
    arc = 6pt,
    colframe = black!50, 
    boxrule = 0pt, 
    bottomrule = 4.5pt ,
    breakable,
}
\usepackage{pifont}

\newcommand{\cmark}{\textcolor{green}{\ding{51}}} 
\newcommand{\xmark}{\textcolor{red}{\ding{55}}}   

\theoremstyle{plain}

\theoremstyle{definition}

\theoremstyle{remark}

\usepackage[textsize=tiny]{todonotes}

\icmltitlerunning{CodeSteer: Symbolic-Augmented Language Models via Code/Text Guidance}

\begin{document}

\twocolumn[
\icmltitle{CodeSteer: Symbolic-Augmented Language Models via Code/Text Guidance}



\icmlsetsymbol{equal}{*}

\begin{icmlauthorlist}
\icmlauthor{Yongchao Chen}{m,ha}
\icmlauthor{Yilun Hao}{m}
\icmlauthor{Yueying Liu}{u}
\icmlauthor{Yang Zhang}{i}
\icmlauthor{Chuchu Fan}{m}
\end{icmlauthorlist}

\icmlaffiliation{m}{Massachusetts Institute of Technology, Boston, MA, USA}
\icmlaffiliation{ha}{Harvard University, Boston, MA, USA}
\icmlaffiliation{i}{MIT-IBM Watson AI Lab, Boston, MA, USA}
\icmlaffiliation{u}{University of Illinois Urbana-Champaign, Urbana, IL, USA}

\icmlcorrespondingauthor{Yongchao Chen}{yongchaochen@fas.harvard.edu}
\icmlcorrespondingauthor{Chuchu Fan}{chuchu@mit.edu}

\icmlkeywords{Machine Learning, ICML}

\vskip 0.3in
]



\printAffiliationsAndNotice{}  


\begin{abstract}
Existing methods fail to effectively steer Large Language Models (LLMs) between textual reasoning and code generation, leaving symbolic computing capabilities underutilized. We introduce CodeSteer, an effective method for guiding LLM code/text generation. We construct a comprehensive benchmark SymBench comprising 37 symbolic tasks with adjustable complexity and also synthesize datasets of 12k multi-turn guidance/generation trajectories and 5.5k guidance comparison pairs. We fine-tune the Llama-3-8B model with a newly designed multi-turn supervised fine-tuning (SFT) and direct preference optimization (DPO). The resulting model, CodeSteerLLM, augmented with the proposed symbolic and self-answer checkers, effectively guides the code/text generation of larger models. Augmenting GPT-4o with CodeSteer raises its average performance score from 53.3 to 86.4, even outperforming the existing best LLM OpenAI o1 (82.7), o1-preview (74.8), and DeepSeek R1 (76.8) across all 37 tasks (28 seen, 9 unseen). Trained for GPT-4o, CodeSteer demonstrates superior generalizability, providing an average 41.8 performance boost on Claude, Mistral, and GPT-3.5. CodeSteer-guided LLMs fully harness symbolic computing to maintain strong performance on highly complex tasks. Models, Datasets, and Codes are available at \url{https://github.com/yongchao98/CodeSteer-v1.0} and \url{https://huggingface.co/yongchao98}.
\end{abstract}

\begin{figure*}[ht]
  \centering
  \includegraphics[width=0.98\linewidth]{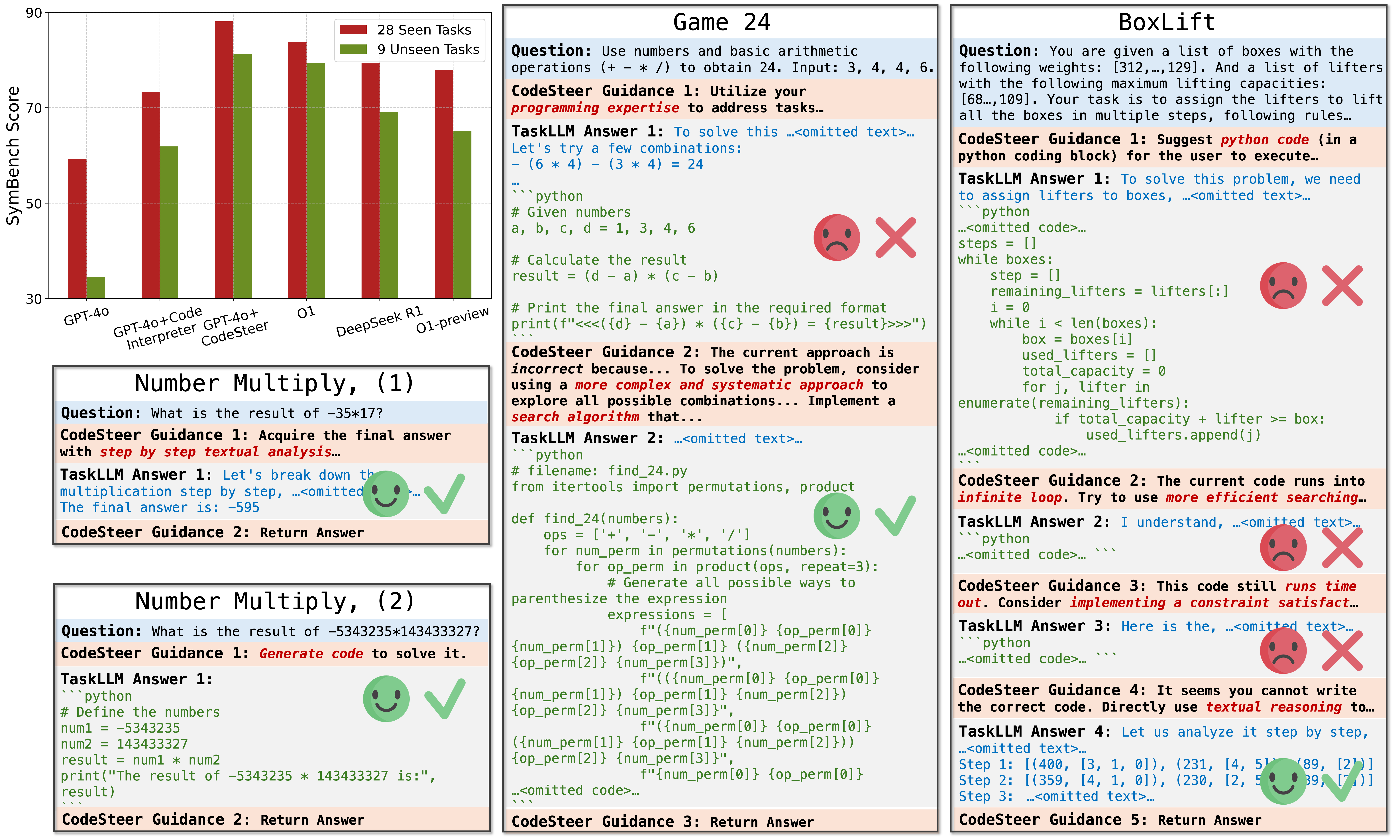}
   \caption{Examples and performance of CodeSteer on guiding LLM code/text generation to integrate symbolic computing. At each interaction with TaskLLM, it reviews current and previous answers, then provides guidance for the next turn. CodeSteer returns final answers when it deems them ready. With CodeSteer, GPT-4o outperforms OpenAI Code Interpreter, o1, and o1-preview models.}
   \label{fig:CodeSteer-intro}
\end{figure*}

\section{Introduction}
\label{Introduction}
While the reasoning and planning capabilities of LLMs have improved significantly~\citep{mixture-of-agents,PROMST,chain-of-code}, they still fail in ostensibly simple tasks \citep{nature-paper-llm-not-reliable}. Crucially, many tasks in existing benchmarks—such as Blocksworld \citep{planbench} and Game 24 \citep{LATS}—can be completely solved with code solutions. Text-based reasoning excels at semantic understanding and commonsense inference but is less suited for exact computation, symbolic manipulation, optimization, and algorithmic processing \citep{llm-cannot-plan-2}. In contrast, symbolic computing via code generation is adept at handling rigorous operations and can easily leverage specialized tools (e.g., equation solvers). In many tasks, prompting LLMs to generate and execute code outperforms purely textual reasoning~\citep{llm+code=commense-learner,code-as-policies,Program-of-thoughts-prompting}.

A key challenge is guiding LLMs to decide when to rely on textual reasoning versus programmatic solutions, given that most input questions lack explicit cues about which approach is best. Recent OpenAI GPT models address this by providing a Code Interpreter module, allowing the model to iteratively generate and execute code, then further reason with the output \citep{gpt-4}. Multi-agent frameworks like AutoGen \citep{autogen} adopt a specialized system prompt to steer LLM for code generation when needed. However, recently \citet{codesteering} finds that all these existing methods struggle to effectively steer between textual reasoning and code generation, failing to fully leverage symbolic computing capabilities.

Our work tries to bridge this gap by developing an assistant framework (CodeSteer) to guide the code/text generation of the LLM solving the task (TaskLLM). By fine-tuning a small model (Llama-3-8B~\citep{llama-3-report}) to be the assistant, we enable large models (GPT-4o~\citep{gpt-4}) to fully leverage symbolic computing via code generation while preserving other capabilities. Recognizing that iterative ``executing and exploring" is the most effective way to solve tasks, we build CodeSteer to generate prompts that guide the TaskLLM through multiple turns of interaction before finalizing answers.

To achieve a comprehensive evaluation, we gather and develop a benchmark with 37 symbolic tasks, referred as SymBench. On SymBench, augmenting GPT-4o with CodeSteer greatly improves its average performance score from 53.3 to 86.4, even outperforming the current leading pure-text model, OpenAI o1 (82.7)~\citep{O1-model} and DeepSeek R1 (76.8)~\citep{deepseek}. Although trained for GPT-4o, CodeSteer shows great generalizability, delivering an average 41.8 performance gain on Claude-3-5-Sonnet, Mistral-Large, and GPT-3.5. By fully leveraging symbolic computing, CodeSteer-guided LLMs maintain strong performance on highly complex tasks even when o1 fails in all testing cases. Our key contributions are:

\textbf{1) Developing and publishing SymBench:} Prior works by~\citet{codesteering} and~\citet{logicgame} gathered and developed 14 and 31 tasks, respectively, targeting challenges in computation, symbolic manipulation, logic, optimization, spatial reasoning, and constrained planning. However, neither study published the complete code for question/solution synthesis or the full datasets. From these 45 tasks, we select 37 that remain challenging for GPT-4o and redevelop their generation code to produce samples with adjustable complexity. We refer to this newly published benchmark as SymBench.

\textbf{2) New methods for dataset construction and model fine-tuning of SFT and DPO:} We fine-tune Llama-3-8B with the synthesized datasets of 12k multi-turn guidance/generation trajectories (SFT) and 5.5k guidance comparison pairs (DPO). Unlike standard multi-step settings, in CodeSteer’s multi-turn guidance, the TaskLLM outputs a complete answer each turn rather than only at the end. Consequently, we introduce novel components to both the dataset construction and training processes for SFT and DPO, such as data synthesis of dynamic guidance adaptation, emphasis on the final two turns in SFT, comparison score design, and efficient answer sampling in DPO. These modifications result in better performance. Both the final CodeSteer model and created datasets will be released.

\textbf{3) Symbolic checker and self-answer checker:} Observing that TaskLLM frequently produces text-like code that hardcodes answers, neglecting efficient symbolic computation, we introduce a Symbolic Checker to help CodeSteerLLM evaluate code complexity and efficiency. Since most reasoning and planning tasks can be better verified with coding, we add a Self-answer Checker for better judgment of answer correctness of CodeSteerLLM. These two new checkers have been proven to significantly improve the efficiency of dataset synthesis and CodeSteerLLM fine-tuning.

\textbf{4) Proposed CodeSteer Outperforms Nine Baselines and o1:} CodeSteer's superior performance highlights the importance of enhancing LLM reasoning and planning with symbolic computing. This also demonstrates the potential for steering large models to generate smarter code and text by leveraging specialized smaller models.

\section{Symbolic Tasks and SymBench}
\textbf{Challenges in Code/Text Choices}\quad For tasks requiring computation, symbolic manipulation, logic, optimization, spatial reasoning, and constrained planning, coding-based symbolic computing is often more effective than text-based approaches. However, \citet{codesteering} found that steering LLM code/text generation poses significant challenges, even in tasks with apparent symbolic characteristics. The main bottlenecks are: 1) Deciding whether code or text is simpler depends on task type, task complexity, and the LLM’s capabilities, which is hard to judge (see Appendix Sec.~\ref{appendix sec: Impacts of task types, task complexity, and LLM capabilities on choices}). 2) LLM-generated code often appears as text-like scripts that merely hard-code answers rather than enabling efficient symbolic computation, echoing the phenomenon described in \citet{LLM-reason-wild} (see Appendix Sec.~\ref{appendix sec: Varied code versions of the same LLM}).

\textbf{SymBench}\quad \citet{codesteering} and \citet{logicgame} collected 14 and 31 tasks with symbolic factors from various benchmarks such as \citet{big-bench-hard,scalable-multi-robot,Tree-of-thought,gsm8k,MATH-dataset}, but their question-generation code and complete datasets remain private. We redevelop the generation code to automatically synthesize questions with adjustable complexity. Our resulting set of 37 tasks covers reasoning, planning, and execution, testing competencies in mathematics, spatial reasoning, logic, order reasoning, optimization, and search. Details and categorization are provided in Appendix Sec.~\ref{appendix sec: SymBench task description} and Table~\ref{Table:SymBench_class}.

\section{CodeSteer Framework}
Fig~\ref{fig:CodeSteer-intro} illustrates how CodeSteer guides the LLM’s code/text generation. At each turn, CodeSteer reviews the TaskLLM’s current answer and the guidance/answer history, then decides whether to offer new guidance or finalize the response. It performs three key functions:\\
\textbf{1) Initial Method Selection}\quad In the first turn, it chooses whether to solve the task with code or text (e.g., use textual reasoning for small-number multiplication, and code for large-number multiplication in the task Number Multiply).\\
\textbf{2) Dynamic Adaptation}\quad In subsequent turns, it refines guidance or switches methods if issues arise (e.g., encouraging more sophisticated symbolic approaches in Game 24, or switching to textual reasoning after multiple incorrect code attempts in BoxLift).\\
\textbf{3) Answer Finalization When Ready}

The main components of CodeSteer are as follows:\\
\textbf{CodeSteerLLM} is the primary model fine-tuned and used to guide TaskLLM in code/text generation. The input prompt formats for the first and subsequent turns are presented in Appendix Sec.~\ref{appendix sec: Prompt for CodeSteerLLM}. To facilitate answer evaluation, CodeSteerLLM is equipped with two checkers—Self-answer and Symbolic—whose design is inspired by the inherent features of symbolic tasks.\\
\textbf{Self-answer Checker} re-queries TaskLLM to generate and execute code for verifying its current answer, then returns the evaluation results and explanations to CodeSteerLLM. Since many symbolic tasks benefit from code-based verification, this approach often provides a more reliable perspective. The prompt format for the Self-answer Checker is provided in Appendix Sec.~\ref{appendix sec: Prompt for Self-answer Checker}.\\
\textbf{Symbolic Checker} is a rule-based script to analyze the generated code for iteration, search, numeric handling, permutations, and combinations, then returns a complexity summary and score. This helps CodeSteerLLM determine whether the code is sufficiently sophisticated for the task at hand. Since TaskLLM often produces text-like code prone to errors, the Symbolic Checker’s complexity assessment aids, but does not solely dictate, CodeSteerLLM’s decisions. Further details on the checking code and prompt are in Appendix Sec.~\ref{appendix sec: Code for Symbolic Checker}.\\
Beyond enhancing CodeSteerLLM's performance, the Self-answer and Symbolic Checkers also streamline dataset synthesis for SFT and DPO fine-tuning, as discussed in the following sections.

\section{Fine-tuning the CodeSteerLLM}

Among the three modules of CodeSteer, the CodeSteerLLM needs to be fine-tuned to perform the complicated task of steering. The fine-tuning is performed on a subset of SymBench. Specifically,
we randomly select 28 of the 37 SymBench tasks, using a distinct set of samples without overlap with the test samples. This setup allows us to evaluate CodeSteer on 28 seen tasks (with different test samples) and on the remaining 9 unseen tasks. 
The fine-tuning consists of two steps. 
We first fine-tune the Llama-3.1-8B model with SFT, then further optimize it using DPO. Both processes are fine-tuned with full parameter on 4*H100 GPUs for 4-10 epochs. The detailed parameter and hardware settings for fine-tuning and inference processes are discussed in Appendix~Sec.~\ref{appendix sec: Parameter and hardware settings of SFT and DPO fine-tuning}.
We synthesize 12k multi-turn guidance/generation trajectories for SFT and 5.5k guidance comparison pairs for DPO. The specific data number for each task is in Appendix~Sec.~\ref{appendix sec: Synthesized dataset number of each task for SFT and DPO}.

\subsection{Multi-turn SFT}
\label{sec:multi-turn SFT}
To generate supervision data for SFT, we prompt the GPT-4o to serve as both the guiding LLM (i.e., the CodeSteerLLM) and the TaskLLM to generate multiple guidance/generate trajectories. We then filter the trajectories keeping only those that produce correct answers.
To improve success rates, CodeSteerLLM’s prompt is more detailed and includes pre-set knowledge or hints. To increase dataset diversity and enable dynamic adaptation of guided thoughts, this prompt also has different versions. For example, we may let GPT-4o choose all guidance styles, or enforce transitions from code to text or text to code. We set the maximum guidance turns to be 5 and return the final answer once that limit is reached.

\textbf{Multi-turn Gradient Cancellation Issue}\quad In multi-turn trajectories, the SFT process incorporates gradients from each turn. This can lead to gradient cancellation in the early turns. For example, in one task, both [code, return answer] and [text, code, return answer] produce correct results, so if both trajectories are used for fine-tuning, the SFT cannot learn that code is the better first step.\\
\textbf{Data Augmentation}\quad To mitigate this issue, we leverage the fact that the final two turns of guidance are most influential, as the TaskLLM produces new answers each turn while earlier turns primarily provide background. Consequently, we augment the SFT dataset by doubling the weights of the final two turns.

\subsection{Multi-turn DPO}
\begin{figure}[!htbp]
  \centering
  \includegraphics[width=0.8\linewidth]{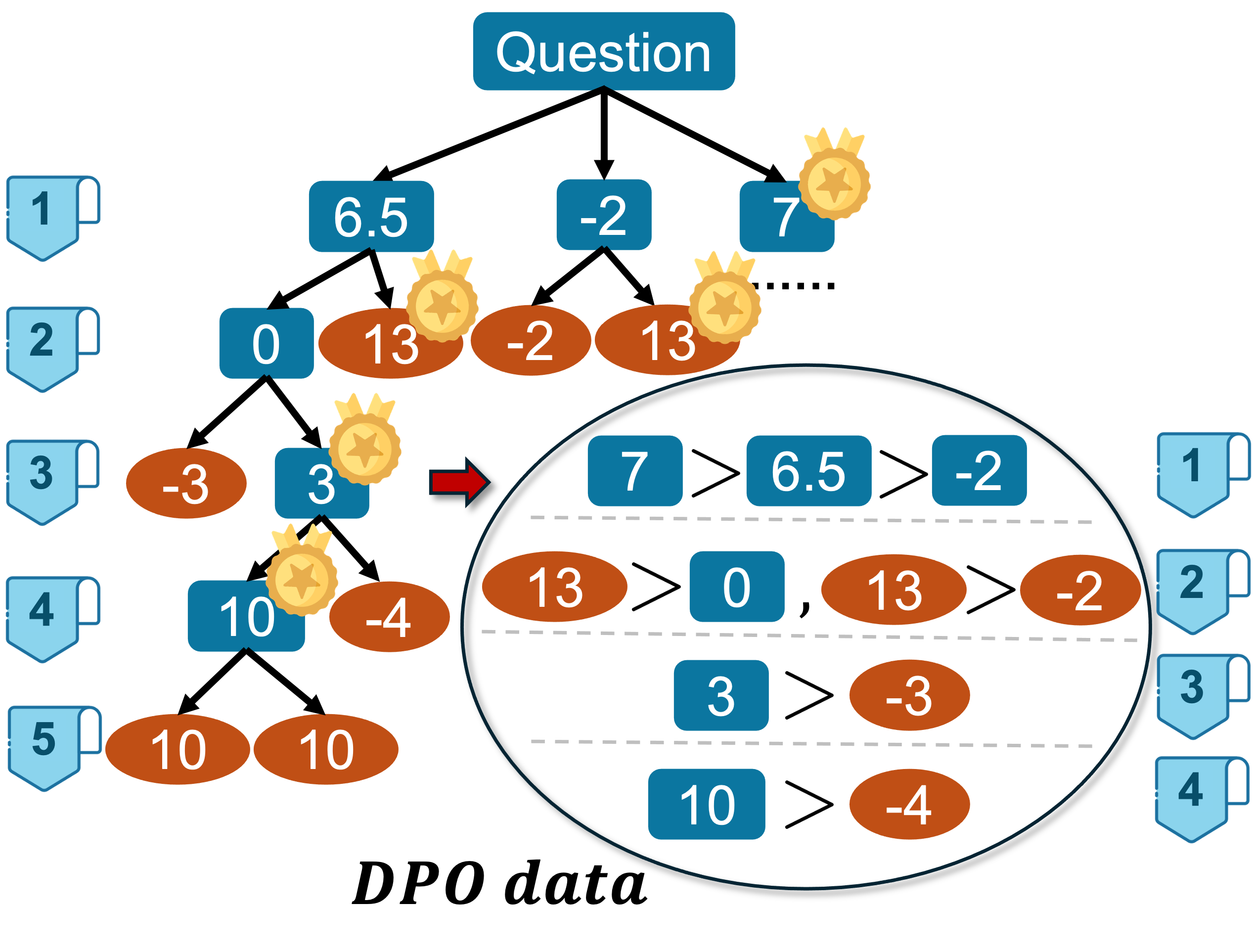}
   \caption{Schematic of multi-turn DPO data sampling: blue squares represent intermediate (non-final) turns, and brown ovals mark finalizing turns. Guidance responses from the same parent node in CodeSteerLLM are compared to generate the DPO data.}
   \label{fig:DPO-data}
\end{figure}

Because many correct trajectories in the SFT dataset are still suboptimal, we need to further fine-tune the CodeSteerLLM on pairs of trajectories labeled with preferences. Here we use rule-based scores to assign preferences. Figure~\ref{fig:DPO-data} illustrates our framework for sampling DPO guidance pairs in a multi-turn setting. The main challenge is sampling and selecting guidance pairs that exhibit clear performance differences across various turns while minimizing the number of samples to conserve resources. We use a tree structure where each node represents a guidance, with a branching factor of 2 or 3. To compare guidance pairs from the same parent node, we calculate their Performance Scores using the following equation:
\begin{equation}
\text{Score}_i =
\begin{cases}
15 - i & \text{ending turn/correct}, \\
- i & \text{ending turn/incorrect}, \\
\frac{1}{|C(i)|} \sum_{j \in C(i)} \text{Score}_j & \text{otherwise}.
\end{cases}
\end{equation}
Here, \( \text{Score}_i \) represents the score for a node at turn \( i \), where \( i \) is the current turn number, and \( C(i) \) is the set of child nodes of node \( i \). If the current turn is the final one, \( \text{Score}_i \) is set to \( 15 - i \) for correct answers and \( -i \) for incorrect ones. This incentivizes CodeSteerLLM to achieve correct answers in the fewest turns possible. For non-final turns, \( \text{Score}_i \) is calculated as the average of its child nodes' scores. This ensures that each non-terminal turn's score reflects the average performance of its potential subsequent actions, i.e., the expectation. 

DPO data is collected from guidance pairs within the same parent node at each level that have a score difference greater than 2. To prevent reward hacking~\citep{reward-hacking}—where CodeSteerLLM might bypass exploration and return incorrect answers quickly (e.g., preferring a score of –2 over –5)—we include only pairs where at least one guidance has a positive score. To obtain diverse guidance answers, we set the inference temperature to 1.5 for the SFT fine-tuned CodeSteerLLM and use three models fine-tuned at different epochs (6, 8, and 10) to compare their guidance responses for the same parent node.

\begin{table*}[!htbp]
\caption{Experimental results on SymBench. Methods with the highest scores are highlighted \textcolor{blue}{blue}.}
\label{table: Full Experimental results on SymBench}
\vskip 0.15in
\begin{center}
\begin{small}
\begin{tabular}{lccccccccccccy}
\toprule
\multicolumn{1}{c}{Methods} & \multicolumn{3}{c}{\textbf{CoT LLMs}} & \multicolumn{6}{c}{\textbf{Training-free Methods}} & \multicolumn{4}{c}{\textbf{Training-based Methods}}\\
\cmidrule(r){2-4} \cmidrule(r){5-10} \cmidrule(r){11-14}
\rotatebox{80}{Task success rate \%} & \rotatebox{80}{o1} & \rotatebox{80}{DeepSeek R1} & \rotatebox{80}{o1-preview} & \rotatebox{80}{Only Question} & \rotatebox{80}{Symbolic Agent} & \rotatebox{80}{All Text + CoT} & \rotatebox{80}{All Code + CoT} & \rotatebox{80}{AutoGen Conca.} & \rotatebox{80}{Code + Text + Sum.1} & \rotatebox{80}{Code + Text + Sum.2} & \rotatebox{80}{Code/Text Choice} & \rotatebox{80}{Code Interpreter} & \rotatebox{80}{GPT-4o + CodeSteer}\\
\midrule
\textbf{Ave. Norm., Seen} & \textbf{83.8} & \textbf{79.3} & \textbf{77.9} & \textbf{59.3} & \textbf{77.0} & \textbf{56.7} & \textbf{71.6} & \textbf{73.2} & \textbf{66.7} & \textbf{65.8} & \textbf{79.7} & \textbf{73.3} & \textbf{\textcolor{blue}{88.1}} \\
\textbf{Ave. Norm., Unseen} & \textbf{79.4} & \textbf{69.1} & \textbf{65.1} & \textbf{34.5} & \textbf{67.9} & \textbf{37.9} & \textbf{63.2} & \textbf{59.5} & \textbf{51.9} & \textbf{51.7} & \textbf{72.1} & \textbf{61.9} & \textbf{\textcolor{blue}{81.3}} \\
\textbf{Ave. Norm., Total} & \textbf{82.7} & \textbf{76.8} & \textbf{74.8} & \textbf{53.3} & \textbf{74.8} & \textbf{52.1} & \textbf{69.6} & \textbf{69.9} & \textbf{63.1} & \textbf{62.4} & \textbf{77.9} & \textbf{70.5} & \textbf{\textcolor{blue}{86.4}}\\
\midrule
\multicolumn{1}{c}{} & \multicolumn{10}{c}{\textbf{Seen Tasks}}\\
Game 24 & 80 & 65 & 78 & 17 & 37 & 23 & 11 & 88 & 33 & 43 & 43 & 18 & \textcolor{blue}{93} \\
Path Plan & 74 & 60 & 56 & 65 & 43 & 44 & \textcolor{blue}{76} & 71 & 66 & 61 & 73 & 54 & 75 \\
BoxLift & \textcolor{blue}{95} & 92 & 85 & 69 & 58 & 56 & 68 & 20 & 65 & 60 & 73 & 49 & 77 \\
BoxNet & 45 & 43 & \textcolor{blue}{54} & 37 & 30 & 30 & 1 & 12 & 23 & 21 & 23 & 37 & 29 \\
Blocksworld & \textcolor{blue}{100} & \textcolor{blue}{100} & 77 & 43 & 60 & 52 & 32 & 50 & 50 & 48 & 44 & 42 & 52 \\
Date Understanding & 87 & 88 & 87 & \textcolor{blue}{90} & 89 & 88 & 72 & 65 & 86 & 84 & 86 & 76 & 87 \\
Web of Lies & \textcolor{blue}{100} & \textcolor{blue}{100} & 98 & 96 & 99 & 86 & 91 & 78 & 77 & 80 & 98 & 94 & 98 \\
Logical Deduction & \textcolor{blue}{100} & 98 & 97 & 89 & 93 & 91 & 83 & 82 & 94 & 90 & 94 & 82 & 92 \\
Navigation & \textcolor{blue}{100} & \textcolor{blue}{100} & \textcolor{blue}{100} & 98 & 93 & 95 & 99 & 91 & 96 & 94 & 92 & 98 & 99 \\
GSM-Hard & 79 & 77 & 71 & 78 & 76 & 80 & \textcolor{blue}{83} & 81 & 81 & 78 & 77 & 79 & 77 \\
MATH Geometry & \textcolor{blue}{94} & 91 & 90 & 76 & 73 & 73 & 74 & 73 & 77 & 76 & 76 & 73 & 75 \\
MATH Count\&Probab. & 96 & \textcolor{blue}{97} & 95 & 89 & 88 & 87 & 88 & 91 & 86 & 88 & 84 & 89 & 93 \\
Logical Equation & \textcolor{blue}{100} & \textcolor{blue}{100} & \textcolor{blue}{100} & 52 & 50 & 52 & 40 & 48 & 30 & 33 & 56 & 71 & 78 \\
New Operator & 44 & 39 & 25 & 42 & 39 & 45 & 39 & 47 & \textcolor{blue}{56} & 38 & 48 & 48 & 40 \\
Pooling & 46 & 40 & 42 & 54 & 46 & \textcolor{blue}{60} & 57 & 55 & 43 & 47 & 40 & 49 & 46 \\
Light Puzzles & \textcolor{blue}{100} & \textcolor{blue}{100} & 92 & 62 & 56 & 56 & 69 & 56 & 92 & 78 & 73 & 95 & 68 \\
Mahjong & 96 & \textcolor{blue}{98} & 93 & 66 & 77 & 73 & 80 & 94 & 72 & 74 & 96 & 64 & 90 \\
Statistical Counting & 25 & 72 & 78 & 34 & 93 & 32 & 95 & 93 & 93 & 86 & 95 & 89 & \textcolor{blue}{97} \\
Matrix Transformation & 87 & \textcolor{blue}{100} & 98 & 94 & 96 & 76 & 97 & 97 & 96 & 92 & 97 & 90 & 98 \\
Logical Puzzle & \textcolor{blue}{88} & 80 & 86 & 48 & 58 & 51 & 41 & 39 & 44 & 50 & 44 & 68 & 70 \\
Cons. Linear Arrange. & 74 & 62 & 81 & 82 & 71 & 84 & 60 & 79 & 72 & 71 & 77 & 72 & \textcolor{blue}{86} \\
Pattern Recognition & \textcolor{blue}{100} & \textcolor{blue}{100} & \textcolor{blue}{100} & 70 & 90 & 44 & 89 & \textcolor{blue}{100} & 56 & 60 & 94 & 100 & 93 \\
String Insertion & 96 & 49 & 72 & 6 & \textcolor{blue}{100} & 8 & 100 & \textcolor{blue}{100} & 67 & 75 & 100 & 89 & \textcolor{blue}{100} \\
Letter Logic Diagram & 50 & \textcolor{blue}{54} & 28 & 2 & 30 & 0 & 12 & 21 & 8 & 9 & 31 & 8 & 45 \\
String Deletion\&Modifi. & 60 & 37 & 34 & 4 & 90 & 0 & 64 & 37 & 51 & 65 & 85 & 49 & \textcolor{blue}{93} \\
String Synthesis & 2 & 0 & 2 & 0 & 20 & 0 & 11 & 0 & 7 & 5 & 16 & 12 & \textcolor{blue}{29} \\
Reversi & 46 & 29 & 28 & 8 & 36 & 15 & 49 & \textcolor{blue}{60} & 20 & 23 & 45 & 23 & 52 \\
Standard Sudoku & 0 & 0 & 0 & 0 & 98 & 0 & 100 & 94 & 12 & 14 & 100 & 100 & \textcolor{blue}{100} \\
\midrule
\multicolumn{1}{c}{} & \multicolumn{10}{c}{\textbf{Unseen Tasks}}\\
Letters & 61 & 52 & 49 & 12 & 91 & 11 & \textcolor{blue}{100} & 93 & 84 & 87 & 89 & 89 & 96 \\
Eight Queen & \textcolor{blue}{84} & 79 & 64 & 8 & 73 & 0 & 35 & 51 & 40 & 45 & 52 & 44 & 78 \\
Number Multiply & 43 & 46 & 28 & 11 & 87 & 8 & \textcolor{blue}{100} & \textcolor{blue}{100} & 68 & 65 & \textcolor{blue}{100} & 75 & 95 \\
Cryptanalysis & 60 & 21 & 49 & 20 & 15 & 24 & 20 & 13 & 16 & 20 & 27 & 0 & 24 \\
String Splitting & \textcolor{blue}{96} & 91 & 90 & 28 & 52 & 25 & 48 & 47 & 37 & 35 & 48 & 43 & 56 \\
Combinatorial Calcul. & 57 & \textcolor{blue}{98} & 35 & 16 & 45 & 60 & 55 & 48 & 70 & 67 & 80 & 57 & 86 \\
Synthesis Decompo. & 57 & \textcolor{blue}{96} & 53 & 52 & 53 & 72 & 71 & 35 & 44 & 38 & 69 & 72 & 66 \\
2048 & 52 & 0 & 37 & 44 & 43 & 40 & 28 & 37 & 25 & 20 & 39 & 49 & \textcolor{blue}{56} \\
Permut. and Combina. & \textcolor{blue}{100} & \textcolor{blue}{100} & \textcolor{blue}{100} & 66 & 89 & 48 & 64 & 60 & 40 & 46 & 80 & 75 & 93 \\
\bottomrule
\end{tabular}
\end{small}
\end{center}
\vskip -0.1in
\end{table*}

\section{Experiments}
\textbf{Experimental settings}\quad We use GPT-4o as the TaskLLM to test 28 seen and 9 unseen tasks, each with 100 samples of varying complexity. The samples for the 28 seen tasks are different from those used to train CodeSteerLLM. Additionally, we evaluate other LLM types to assess CodeSteer’s generalizability.

We compare CodeSteer to six training-free and three training-based baselines, with methods 1, 3–6, and 9 originally proposed in~\citet{codesteering}.\\
\textbf{Training-free Baselines}\quad 1) No extra modifications but only input the original question (\textbf{Only Question}); 2) Our framework in Sec.~\ref{sec:multi-turn SFT} to synthesize SFT dataset, where GPT-4o works as CodeSteerLLM with extra hints (\textbf{Symbolic Agent}); 3) Prompting LLMs to answer with only text with CoT (\textbf{All Text + CoT}); 4) Prompting LLMs to first analyze the question with CoT and then output the code answer (\textbf{All Code + CoT}); 5) Concatenating the input question with AutoGen's original system prompt in Appendix Section~\ref{appendix sec: System prompt of AutoGen} (\textbf{AutoGen Conca.}); 6) Implement a multi-agent framework that first queries LLMs to answer the question with All Text + CoT and All Code + CoT methods, respectively. Then the final solution is obtained by combining and summarizing both versions of the answers by the same LLM but prompted differently (\textbf{Code + Text + Sum.1}).\\
\textbf{Training-based Baselines}\quad 7) Fine-tune Llama-3.1-8B as a summarizer based on the Code + Text + Sum.1 method using SFT on correct summary data (\textbf{Code + Text + Sum.2}); 8) We fine-tune Llama-3.1-8B as a one-step evaluator to choose between text or code generation (\textbf{Code/Text Choice}); 9) OpenAI GPT Code Interpreter with the original input question (\textbf{Code Interpreter}). Method 7 and 8 are fine-tuned on the same data number and task types as CodeSteer.\\
\textbf{Comparison with CoT LLMs}\quad We also compare with the current best models: OpenAI o1 and o1-preview~\citep{O1-model} and DeepSeek R1~\citep{deepseek}. These models enhance reasoning and planning by using textual search, reflection, and exploration during answer generation. However, our analysis shows that these CoT LLMs have not yet integrated code-based symbolic computing to further improve their performance.

\textbf{Evaluations}\quad Answers are evaluated using predefined rules, with GPT-4o assisting in adjusting formats as needed. Beyond the Code Interpreter method, some approaches have the LLM output code as the final answer. We extract and execute this code using predefined algorithms to obtain the final result or facilitate further reasoning. To prevent infinite loops, code execution is limited to 30 seconds. If this limit is exceeded, the task is marked as failed or returns errors for subsequent turns. We utilize success rate as the metric for each task. To compare each method, we calculate the Average Normalized Score over all the tested tasks by the following equation:
\begin{equation}
\text{AveNorm}_j = \frac{1}{N} \sum_{i=1}^{N} \frac{s_{ij}}{\text{max}(s_i)}
\end{equation}
where \( \text{AveNorm}_j \) is the Average Normalized Score for method \( j \), \( s_{ij} \) is the score of method \( j \) for task \( i \), \( \text{max}(s_i) \) is the maximum score for task \( i \), \( N \) is the total number of tasks. This equation normalizes each score relative to the maximum score in the respective task, and then averages the normalized scores over all tasks. We use the normalized metric to better compare the relative performance among methods and prevent any single task from disproportionately influencing the overall evaluation. As shown in Appendix Sec~\ref{appendix sec: Score comparison of different methods without normalization across tasks}, to ensure the robustness of our conclusions against changes in the evaluation metric, we recalculate the average score without normalization. From the result in Appendix Table~\ref{table:avg-score without normalization}, we can observe that for both seen and unseen tasks, our method can still outperform all main baselines obviously, even more on unseen tasks.

Apart from the task performance, in later sections we also discuss the costs of token lengths and runtime for each method.

\subsection{Overall Better Performance}
\begin{figure}[!htbp]
  \centering
  \includegraphics[width=0.7\linewidth]{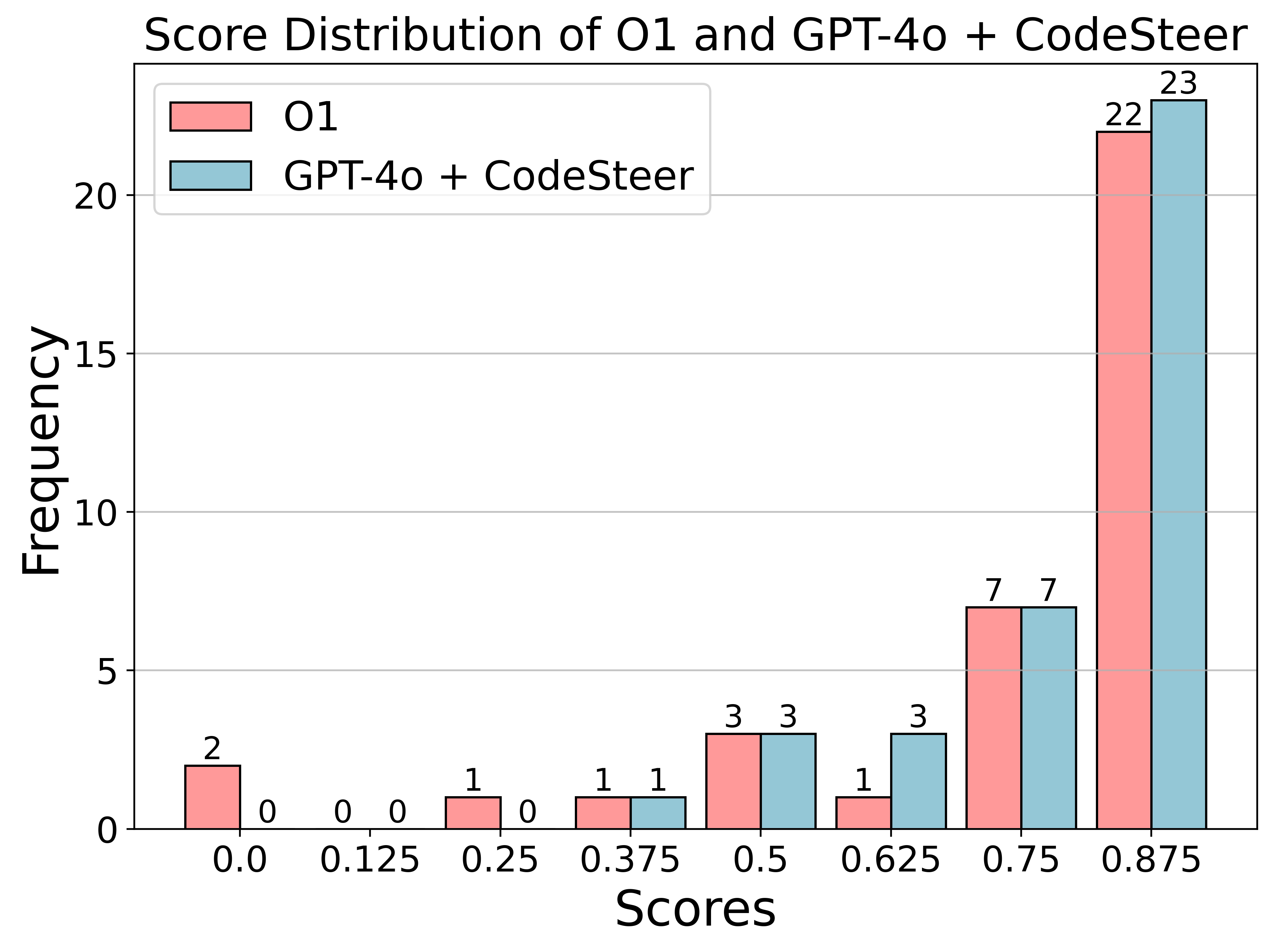}
   \caption{Normalized score distribution of CodeSteer+GPT-4o and o1 in 37 SymBench tasks.}
   \label{fig:score_distribution}
\end{figure}

Table~\ref{table: Full Experimental results on SymBench} presents the full results of all methods on SymBench, including individual task scores and the Average Normalized Score. The key findings are:\\
\textbf{1)}\quad CodeSteer maintains similar relative performance on seen and unseen tasks, indicating no overfitting.\\
\textbf{2)}\quad Augmenting GPT-4o with CodeSteer significantly boosts its performance, raising the Ave. Norm. Total Score from 53.3 to 86.4—outperforming all 9 baselines (best baseline: Code/Text Choice at 77.9).\\
\textbf{3)}\quad GPT-4o + CodeSteer surpasses o1 (82.7), R1 (76.8), and o1-preview (74.8), highlighting the importance of integrating symbolic computing into LLMs. Figure~\ref{fig:score_distribution} compares the score distribution of GPT-4o + CodeSteer and o1, showing that CodeSteer reduces instances of extremely low scores (near 0), demonstrating its robustness to varied tasks.\\
\textbf{4)}\quad Compared to other training-based methods (Code + Text + Sum.2 and Code/Text Choice) with the same data number and tasks, CodeSteer’s better performance validates the framework's effectiveness (further discussed in Sec.~\ref{sec: Ablation Studies}).

\begin{table*}[!htbp]
\caption{Experimental results of Claude-3-5-sonnet-20241022, Mistral-Large, and GPT-3.5 with or without augmented CodeSteer. Methods with the higher scores of the same model are highlighted \textcolor{blue}{blue}.}
\label{table: Claude-mixtral-gpt3}
\vskip 0.15in
\begin{center}
\begin{small}
\begin{tabular}{lyyggaa}
\toprule
Methods & Claude & Claude + CodeSteer & Mistral & Mistral + CodeSteer & GPT-3.5 & GPT-3.5 + CodeSteer\\
\hline
Combinatorial Calcu. & 48 & 66 & 25 & 34 & 12 & 29\\
Eight Queen & 4 & 87 & 60 & 41 & 0 & 16\\
Reversi & 0 & 45 & 0 & 33 & 0 & 32\\
Cons. Linear Arran. & 73 & 90 & 47 & 48 & 25 & 9\\
Standard Sudoku & 0 & 100 & 0 & 100 & 0 & 95\\
\hline
\textbf{Ave. Norm. Score} & \textbf{29.1} & \textbf{\textcolor{blue}{92.0}} & \textbf{31.0} & \textbf{\textcolor{blue}{59.8}} & \textbf{8.6} & \textbf{\textcolor{blue}{42.3}}\\
\bottomrule
\end{tabular}
\end{small}
\end{center}
\vskip -0.1in
\end{table*}

\subsection{Scalability and Generalizability}
\begin{figure}[!htbp]
  \centering
  \includegraphics[width=0.98\linewidth]{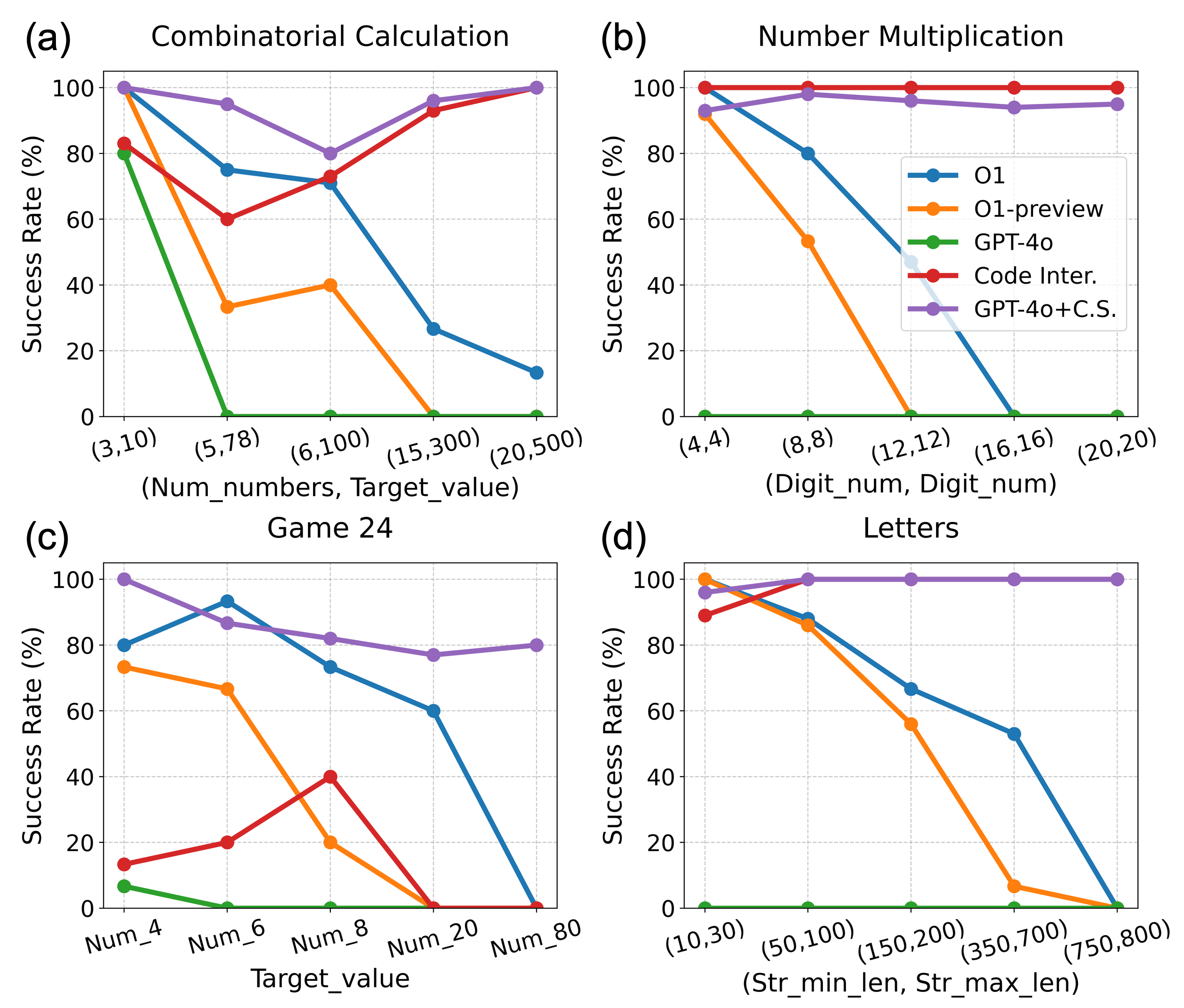}
   \caption{Method performance across four representative tasks as task complexity increases from left to right on the x-axis controlled by value scales. C.S. and Inter. represent CodeSteer and Interpreter.}
   \label{fig:Varied-complexity}
\end{figure}

To assess the impact of symbolic computing, Fig.~\ref{fig:Varied-complexity} tracks the performance of five methods across four tasks of increasing complexity. As critical task-specific properties escalate, o1, o1-preview, and GPT-4o fail in highly complex cases, while symbolic-augmented methods (CodeSteer, Code Interpreter) sustain performance. Notably, CodeSteer proves more robust across tasks than Code Interpreter.

In our study, CodeSteerLLM is fine-tuned on synthesized datasets where TaskLLM is always GPT-4o. To assess its transferability and generalizability, we test it with three popular models: Claude-3-5-Sonnet, Mistral-Large, and GPT-3.5-Turbo. We evaluate them on five representative tasks based on GPT-4o’s results in Table~\ref{table: Full Experimental results on SymBench}: two where text outperforms code and three where code is superior. CodeSteer has shown apparent effects when guiding GPT-4o on these tasks. The results in Table~\ref{table: Claude-mixtral-gpt3} confirm that CodeSteer generalizes well across other LLMs types. This is expected, as its core mechanisms—code/text guidance and dynamic adaptation—are essential to all general-purpose LLMs. Notably, we observe that CodeSteer is particularly effective when applied to stronger LLMs, such as Claude. This is likely because more powerful models possess superior self-reflection capabilities and can generate complex code with greater precision. Thus, they benefit more from CodeSteer’s additional structured guidance, unlocking their full potential.

\subsection{Cost of Tokens and Runtime}
\begin{figure}[!htbp]
  \centering
  \includegraphics[width=1.0\linewidth]{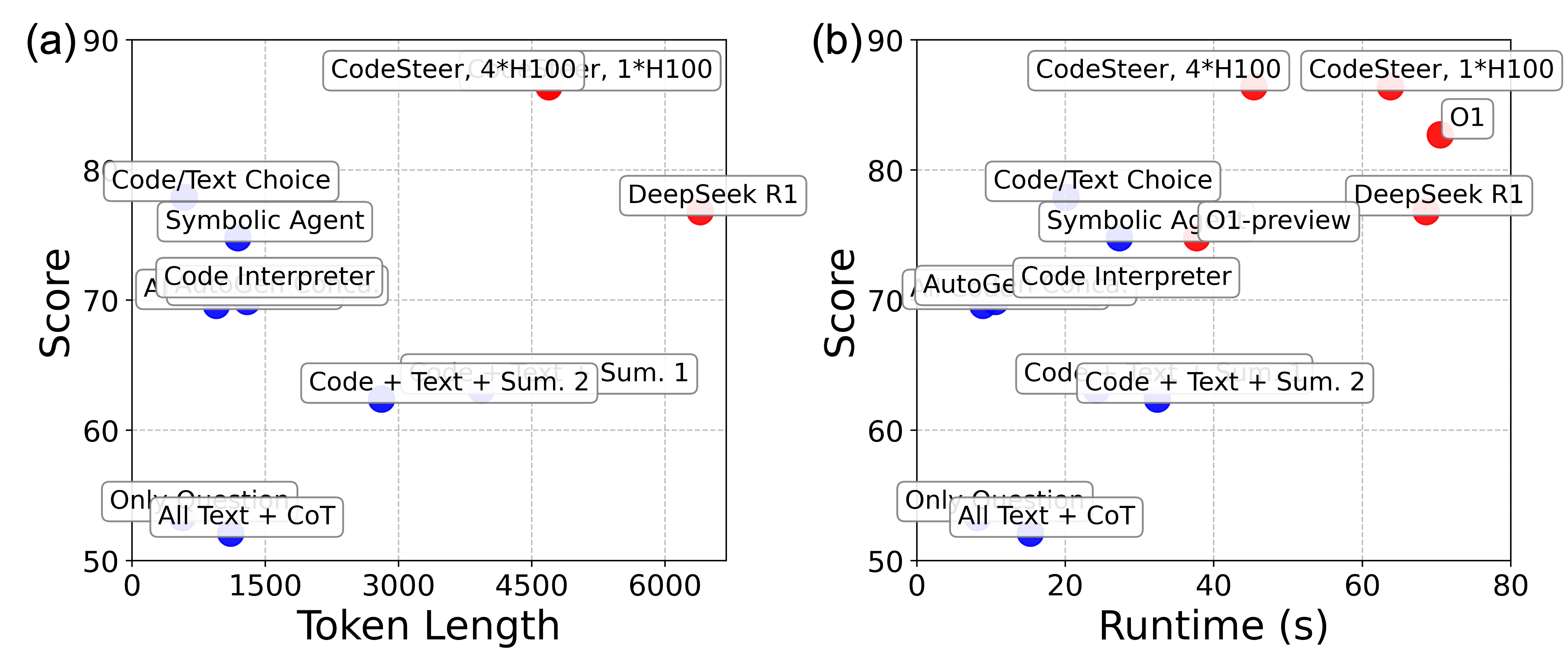}
   \caption{Score vs. token and runtime costs for each method, highlighting CodeSteer, R1, o1, and o1-preview in red. We display CodeSteer results separately for inferences using single or four H100 GPUs. Specific values are in Table~\ref{table: Score-cost table for each method}.}
   \label{fig:Cost-token-runtime}
\end{figure}

Figure~\ref{fig:Cost-token-runtime} shows Score versus Token Length (including input and output tokens) and Score versus Runtime (covering both LLM inference and code execution) for all methods. Complete data is provided in Appendix Table~\ref{table: Score-cost table for each method}. Token counts include only those used by TaskLLM, excluding small and open-source models fine-tuned on Llama-3.1-8B. For the o1 and o1-preview models, only runtime is plotted since their thinking chains are unavailable. While achieving superior performance, CodeSteer uses more tokens than baseline methods due to its multi-turn generations. Most of these tokens are consumed by multiple interaction turns that ultimately fail. CoT LLM R1 consumes more tokens than CodeSteer due to the inefficient textual iteration.

In terms of runtime, CodeSteer is faster than o1 and R1 while delivering better performance. Additionally, since most of CodeSteer’s runtime comes from the inference of the 8B CodeSteerLLM on our workstation, hardware and system optimizations can significantly reduce it. For example, running CodeSteerLLM on four H100 GPUs instead of one decreases the average runtime from 63.8 to 45.4 seconds. CoT LLMs consume excessive runtime and tokens due to their extensive and often redundant reasoning chains. Textual iteration is inherently inefficient for search. Appendix~Sec.~\ref{Appendix sec:Example of GPT-4o Text Answer in Game 24} shows examples of text answers of R1 and GPT-4o, in which both models attempt to find the correct equation for the Game 24 task by listing all possible combinations, leading to uncontrolled iterations and endless generation. This highlights the importance of symbolic computing through code generation.

\subsection{o1 + CodeSteer}
Here we test whether CodeSteer can improve the performance of reasoning models like o1. As shown in Table~\ref{table:o1+CodeSteer}, the performance of o1 will improve notably when augmented with CodeSteer on 5 randomly chosen unseen tasks, further verifying the effectiveness of CodeSteer.

\begin{table*}[!htbp]
\caption{Per-task success rate (\%) for the text-only baseline (o1) versus o1 augmented with CodeSteer.  For each task, the higher score is highlighted in {\color{blue}{blue}}.}
\label{table:o1+CodeSteer}
\begin{center}
\begin{small}
\begin{tabular}{lccccc}
\toprule
\textbf{Task success rate \%} & \textbf{Cryptanalysis} & \textbf{Synthesis Decomposition} & \textbf{2048} & \textbf{Eight Queens} & \textbf{Combinatorial Calculation} \\
\midrule
o1 & \textbf{60} & \textbf{57} & \textbf{52} & \textbf{84} & \textbf{57} \\
o1 + CodeSteer & \textbf{\textcolor{blue}{73}} & \textbf{\textcolor{blue}{94}} & \textbf{\textcolor{blue}{72}} & \textbf{\textcolor{blue}{97}} & \textbf{\textcolor{blue}{95}} \\
\bottomrule
\end{tabular}
\end{small}
\end{center}
\end{table*}

\subsection{Comparison with Heuristic-based Methods}
To better evaluate CodeSteer and provide deeper comparisons, we include three prompt-based baselines. \textbf{Few-Shot}: Uses five example-based prompts to guide the TaskLLM in mimicking the `code/text' switching reasoning process. \textbf{Code-First-Rule}: A rule-based approach where the TaskLLM is prompted to use code for the first three rounds (with increasing complexity) and then switch to text-based reasoning. \textbf{Code-First-Agent}: Employs GPT-4o as the CodeSteerLLM to guide the TaskLLM using the same code-first-then-text strategy as in Code-First-Rule. As shown in Appendix Table~\ref{table:heuristic-based methods}, the three prompt-based methods perform significantly worse than CodeSteer, underscoring the effectiveness of training with our synthesized data. Upon analyzing failure cases, we identify two main reasons:

\textbf{1)}\quad CodeSteerLLM’s guidance often includes problem-specific coding knowledge (e.g., suggesting A* or DFS) and how to formalize the problem, which purely prompt-based methods struggle to capture.

\textbf{2)}\quad Switching between code and text can be advantageous, as later code generations can build on insights from prior textual reasoning. For instance, in Path Plan, a text-generated trajectory may be partially correct; subsequent code can refine it directly, reducing the search space.

\section{Ablation Studies}
\label{sec: Ablation Studies}
\begin{table*}[!htbp]
\caption{Ablation studies on CodeSteer. WO DPO: CodeSteer with SFT but without DPO fine-tuning. WO DPO WO Data Augment: Same as WO DPO, but without data augmentation in the last two turns. Agent represents the Symbolic Agent.}
\label{table:ablation-study}
\begin{center}
\begin{small}
\begin{tabular}{lcccccccc}
\toprule
Methods & 1.Code & 2.WO & 3.WO DPO & 4.WO& 5.WO & 6. & 7.Agent WO & 8.Agent WO\\
& Steer& DPO & WO Data & Symbolic & Self-answer & Agent& Symbolic & Self-answer \\
Task success rate \% & & & Augment. & Checker & Checker & & Checker & Checker \\
\midrule
\textbf{Ave. Norm., Seen} & \textbf{\textcolor{blue}{88.1}} & \textbf{80.0} & \textbf{79.7} & \textbf{80.1} & \textbf{78.5} & \textbf{77.0} & \textbf{71.9} & \textbf{70.1} \\
\textbf{Ave. Norm., Unseen} & \textbf{\textcolor{blue}{81.3}} & \textbf{76.2} & \textbf{70.9} & \textbf{68.6} & \textbf{64.2} & \textbf{67.9} & \textbf{62.0} & \textbf{57.4} \\
\textbf{Ave. Norm., Total} & \textbf{\textcolor{blue}{86.4}} & \textbf{79.1} & \textbf{77.6} & \textbf{77.3} & \textbf{75.0} & \textbf{74.8} & \textbf{69.5} & \textbf{67.0} \\
\bottomrule
\end{tabular}
\end{small}
\end{center}
\end{table*}

The CodeSteer framework comprises SFT and DPO dataset synthesis, CodeSteerLLM fine-tuning, a symbolic checker, and a self-answer checker. Here we do the ablation studies on these components and their related modifications. The added experimental results are shown in Table~\ref{table:ablation-study} with the whole result table of 37 SymBench tasks in Append~Sec.~\ref{appendix sec: Full experimental results of ablation studies}.

\textbf{DPO Effects}\quad In Table~\ref{table:ablation-study}, 1.CodeSteer outperforms 2.WO DPO, showing the effectiveness of the DPO process.

\textbf{SFT Data Augmentation}\quad As discussed in Sec.~\ref{sec:multi-turn SFT}, we do the data augmentation of the last two turns in each trajectory to prevent multi-turn gradient cancellation. In Table~\ref{table:ablation-study}, 2.WO DPO achieves higher score than 3.WO DPO WO Data Augment., which means this extra attention on the last two turns does enhance the SFT process.

\textbf{Symbolic and Self-answer Checkers}\quad We evaluate the effects of the Symbolic and Self-answer Checker in two parts: \textbf{1) Dataset Synthesis Efficiency:} Comparing Group 6 with Groups 7 and 8 in Table~\ref{table:ablation-study} shows that integrating these two checkers increases the Symbolic Agent's success rates, thereby enhancing the efficiency of the dataset synthesis process. \textbf{2) CodeSteer Performance:} Comparing Group 1 with Groups 4 and 5 demonstrates that augmenting with these two checkers improves CodeSteer’s final performance.

\textbf{Multi-turn Guidance}\quad CodeSteer uses a multi-turn interaction strategy with TaskLLM. In contrast, the Code/Text Choice method in Table~\ref{table: Full Experimental results on SymBench} relies on single-step guidance and performs worse than CodeSteer. This demonstrates that the multi-turn design enhances guidance effectiveness, aligning with the common intuition that the best methods for many tasks emerge from iterative ``executing and exploring" processes accompanied with dynamic adaptation.

\textbf{Guide Not Summarizer}\quad CodeSteer primarily serves as the guidance generator for TaskLLM rather than directly generating answers, summarizing, or selecting among multiple answers. This design choice accounts for the limitations of the open-source LLM we use compared to the more capable closed-source LLM that supports TaskLLM. By focusing on guidance, CodeSteer reduces task complexity and data space requirements. The Code + Text + Sum.2 approach in Table~\ref{table: Full Experimental results on SymBench} attempts to fine-tune an answer summarizer using the same data volume but fails, highlighting that summarization imposes a significant burden on Llama-3.1-8B due to the unique characteristics of each task.

\section{Related Work}
\label{sec: Related Work}
\textbf{Code Generation and Symbolic Computing in LLM Tasks}\quad LLMs are widely used for general agent tasks, such as interacting with softwares and websites \citep{webarena,travelplanner,llmfp,crab}, planning robot actions \citep{scalable-multi-robot,saycan}, and inferring with logic~\citep{big-bench-hard}. Literally, many test tasks in previous works can be solved with direct coding~\citep{meta-prompting,pal}. Some recent works also further extend the applications of coding into tasks involving commonsense reasoning and semantic analysis~\citep{chain-of-code,weir2024learning}. Most of previous works mainly utilize text~\citep{Tree-of-thought,saycan,text2motion} or code~\citep{code-as-policies,codeplan-code-use-llm,code-based-self-verify} as the only output modality. \citet{codesteering} highlights the importance of smartly switching between code and text generation in LLMs but notes current methods have clear drawbacks.\\
\textbf{LLM Self-reflection and CoT Models}\quad LLM-generated feedback via self-evaluation can improve performance on a variety of tasks \citep{yang2022re3, welleck2022generating, madaan2023self}. The OpenAI o1~\citep{O1-model} and DeepSeek R1~\citep{deepseek} models demonstrate the potential of agentic LLMs that use Chain-of-Thought (CoT) text generation to explore and self-reflect, enhancing reasoning and planning. However, they lack symbolic computing and code generation capabilities, leading to weaker performance on complex symbolic tasks and consuming substantial tokens and time~\citep{overthink-o1,Ai_2025}.\\
\textbf{LLM Fine-tuning with Multi-step SFT and DPO}\quad SFT~\citep{SFT-self-play} and DPO~\citep{DPO} are extensively implemented for LLM fine-tuning. To enhance LLM's capability in multi-step agent tasks, these methods are further modified with multi-step goals and rewards~\citep{multi-turn-RLHF,VLM-RL-multi-Turn,CPO}. LLM self-generated data have become increasingly important for model improvement when combined with search algorithms and rejection sampling~\citep{LATS,rstar-math}.

\section{Discussion}
Our work underlines the significance of augmenting LLM reasoning and planning capabilities with symbolic computing and shows great potentials of steering large models for smarter code/text generation with specialized small models. We introduce novel modifications to dataset synthesis and fine-tuning (SFT/DPO) to support a multi-turn guidance framework, which has proven effective. Unlike CoT LLMs like OpenAI o1 and DeepSeek R1, which rely solely on textual reasoning for exploration, symbolic computing offers greater efficiency, robustness, and scalability. Since coding is a core LLM capability, generating symbolic tools via code writing preserves generalization across tasks.

\textbf{Limitations}\quad We note that CodeSteer encounters failures under the following conditions, insisting the further research to overcome these bottlenecks.

\textbf{1)}\quad Insufficient Capability of TaskLLM: In some cases, the capabilities of the TaskLLM—whether through coding or textual reasoning—are not sufficient to solve the given problem.

\textbf{2)}\quad Suboptimal Code Generation: The generated code may not use the most efficient method, which can lead to timeouts. For example, as shown in Fig.~\ref{fig:Varied-complexity}c, CodeSteer’s success rate decreases when the target values increase, due to the exponential growth in search complexity.

\textbf{3)}\quad Lack of Robustness to Task Complexity: CodeSteer is not yet robust enough across tasks with varying complexity. As shown in Fig.~\ref{fig:Varied-complexity}a, performance drops in medium-complexity samples. In these cases, CodeSteer sometimes selects textual reasoning over coding and ends up producing incorrect answers.

\section*{Acknowledgments}
This work was supported by ONR under Award N00014-22-1-2478 and MIT-IBM Watson AI Lab. However, this article solely reflects the opinions and conclusions of its authors.

\section*{Impact Statement}
This paper aims to advance the field of Foundation Models. Steering the generation from language models has the great potential to improve safety and performance to better align with human preferences. Any such work is inherently a double-edged sword; the same techniques used to generate samples from a harmless distribution of text could, with a single sign change, be repurposed for generating samples from a harmful distribution of text. Our method improves language model capability by integrating symbolic computing, which may also be misused for harmful purposes.

Overall, we believe the potential positive social benefits of our work in evaluation and steering language model output towards desired target distributions outweigh the potential negatives stemming primarily from misuse.

\nocite{langley00}

\bibliography{example_paper}
\bibliographystyle{icml2025}


\newpage
\appendix
\onecolumn

\section*{Appendices: CodeSteer: Symbolic-Augmented Language Models via Code/Text Guidance}
\addcontentsline{toc}{section}{Contents of Appendices}

\noindent
A. Impacts of task types, task complexities, and LLM capabilities on code/text choices \dotfill \pageref{appendix sec: Impacts of task types, task complexity, and LLM capabilities on choices}
\medskip

B. Varied code versions of the same LLM \dotfill \pageref{appendix sec: Varied code versions of the same LLM}
\medskip

C. Description of SymBench tasks \dotfill \pageref{appendix sec: SymBench task description}
\medskip

D. Prompt for CodeSteerLLM \dotfill \pageref{appendix sec: Prompt for CodeSteerLLM}
\medskip

E. Prompt for Self-answer Checker \dotfill \pageref{appendix sec: Prompt for Self-answer Checker}
\medskip

F. Code for Symbolic Checker \dotfill \pageref{appendix sec: Code for Symbolic Checker}
\medskip

G. Synthesized dataset number of each task for SFT and DPO \dotfill \pageref{appendix sec: Synthesized dataset number of each task for SFT and DPO}
\medskip

H. Parameter and hardware settings of SFT/DPO fine-tuning and inference processes \dotfill \pageref{appendix sec: Parameter and hardware settings of SFT and DPO fine-tuning}
\medskip

I. Score comparison of different methods without normalization across tasks \dotfill \pageref{appendix sec: Score comparison of different methods without normalization across tasks}
\medskip

J. Score-cost table for each method \dotfill \pageref{Appendix sec:Score-cost table for each method}
\medskip

K. Comparison with heuristic-based methods \dotfill \pageref{Appendix sec:Score table for comparison with heuristic-based methods}
\medskip

L. Example text answer of DeepSeek R1 and GPT-4o in Game 24 \dotfill \pageref{Appendix sec:Example of GPT-4o Text Answer in Game 24}
\medskip

M. Full experimental results of ablation studies \dotfill \pageref{appendix sec: Full experimental results of ablation studies}
\medskip

N. System prompt of AutoGen \dotfill \pageref{appendix sec: System prompt of AutoGen}

\newpage
\section{Impacts of task types, task complexities, and LLM capabilities on code/text choices}
\label{appendix sec: Impacts of task types, task complexity, and LLM capabilities on choices}
\begin{figure*}[!htbp]
  \centering
  \includegraphics[width=0.95\linewidth]{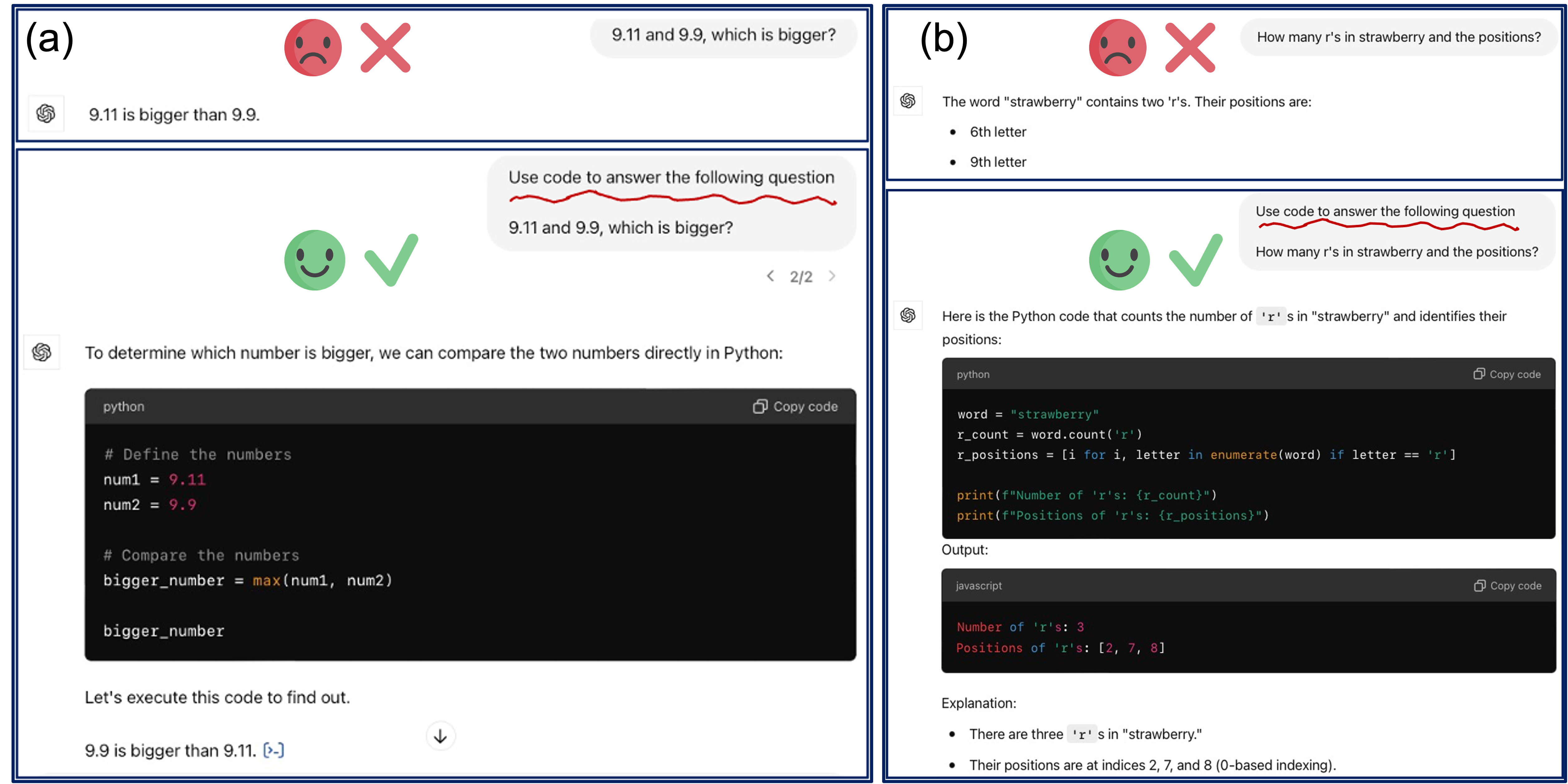}
   \caption{The cases that GPT-4o makes simple mistakes by direct textual reasoning but can reliably solve the problem with prompted to use code.}
   \label{fig:GPT4o-makes-simple-mistakes-in-number-letter}
\end{figure*}

\begin{figure*}[!htbp]
  \centering
   \includegraphics[width=0.85\linewidth]{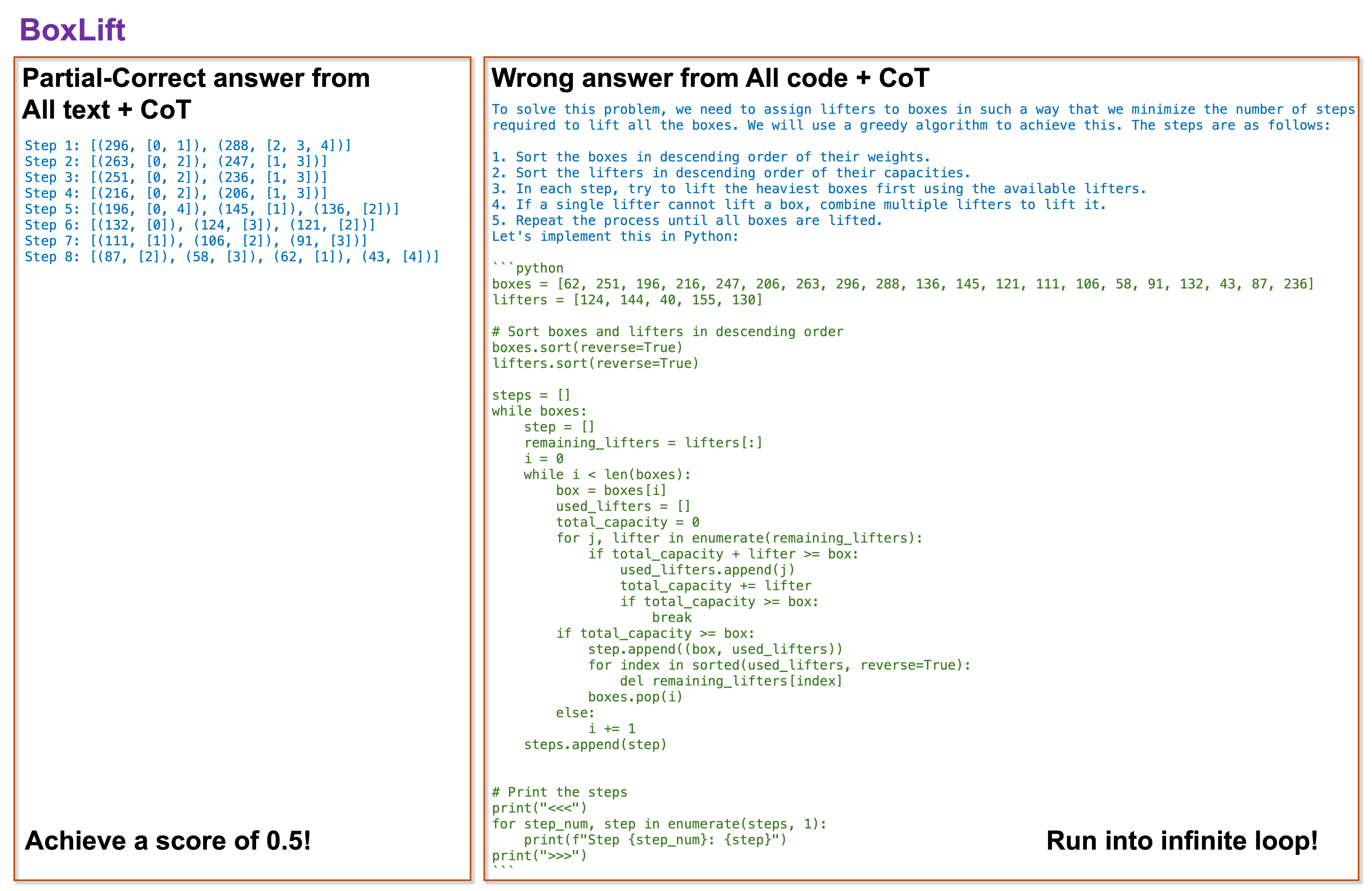}
   \caption{Representative answers of BoxLift task. The left figure is the partially correct answer of GPT-4o with All Text + CoT method. The right figure is the wrong code answer from All Code + CoT method. The text and code parts are colored in blue and green, respectively. The All Code + CoT method generates the wrong code that runs into an infinite loop.}
   \label{fig:BoxLift-answer-example}
\end{figure*}

The phenomenon and challenges of steering LLM code/text generation are first proposed by~
\citet{codesteering}. Here we discuss these phenomenon in details for the motivation of our work. Fig~\ref{fig:GPT4o-makes-simple-mistakes-in-number-letter} presents two typical examples of the recently popular topics of '9.11' and '9.9' numerical comparison and 'r' letter count in 'strawberry', that the ChatGPT of GPT-4o makes mistakes by direct textual reasoning but easily solves the problem after prompted to use code. Meanwhile, Fig~\ref{fig:BoxLift-answer-example} displays the example that GPT-4o makes mistakes to solve the question by code generation but partially solve the question by textual reasoning. The above two examples show that whether code or text is simpler highly depends on the task types and LLM own capabilities and characteristics.

\begin{figure*}[!htbp]
  \centering
   \includegraphics[width=0.8\linewidth]{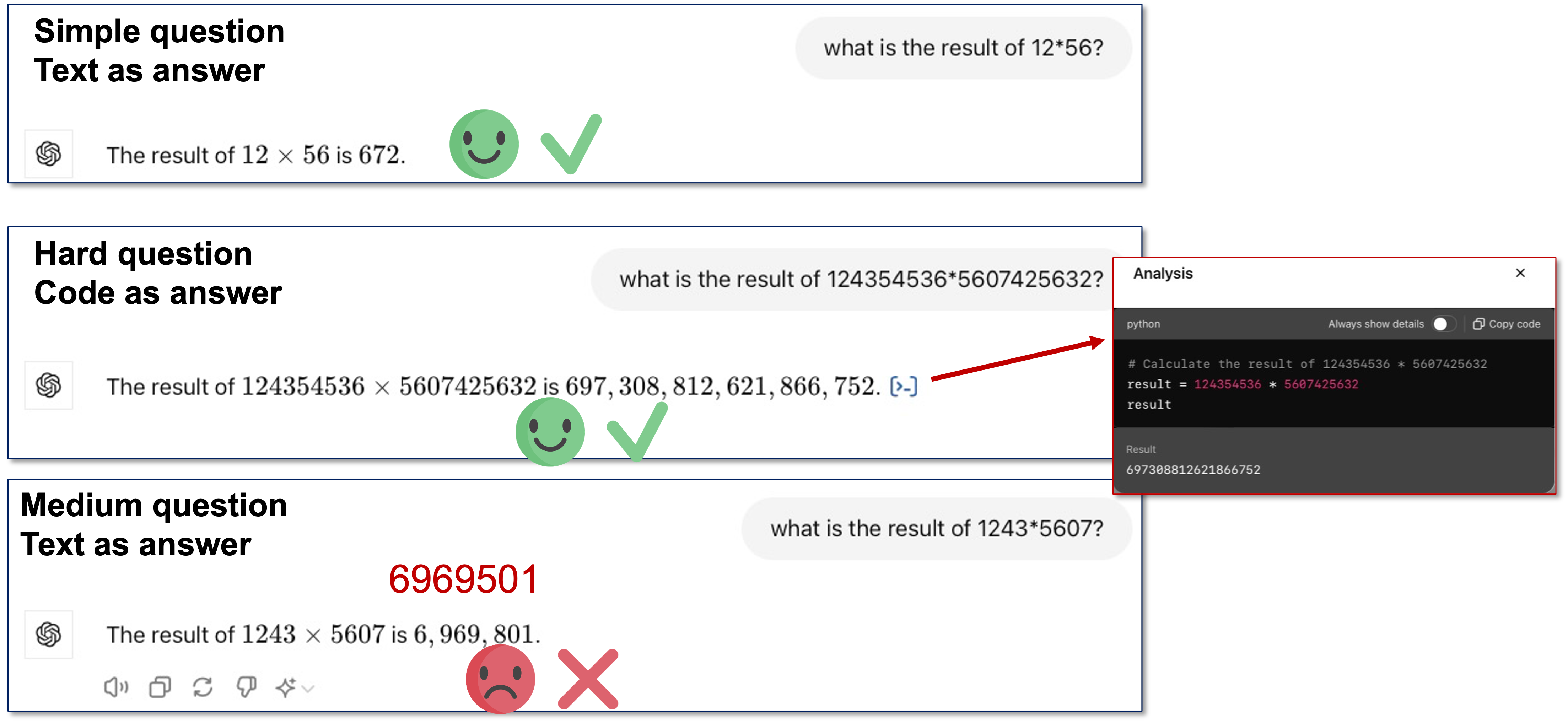}
   \caption{GPT-4o Code Interpreter tends to handle simple Number Multiplying tasks with text and complex tasks with code. However, it often fails with medium-difficulty questions, where it is overconfident and chooses not to use code when needed.}
   \label{fig:Evolution-with-complexity-number-multiply}
\end{figure*}

The OpenAI GPT-4o Code Interpreter is trained to steer LLM code/text generation. However, the study of~\citet{codesteering} finds many limitations of this method. In Fig~\ref{fig:Evolution-with-complexity-number-multiply}, they observe an intriguing property of GPT Code Interpreter: its decision to use code depends on the complexity of the task, as shown in Fig~\ref{fig:Evolution-with-complexity-number-multiply}. GPT-4o Code Interpreter chooses to handle simple Number Multiplying questions with text and complex questions with code, resulting in correct answers. However, it fails in medium-difficulty questions since it tends to be overconfident and chooses to answer the question via textual reasoning, which sometimes is wrong. Hence, whether to implement symbolic computing depends on task complexities even for the same type of the task.

\newpage
\section{Varied code versions of the same LLM}
\label{appendix sec: Varied code versions of the same LLM}
\begin{figure*}[!htbp]
  \centering
   \includegraphics[width=0.95\linewidth]{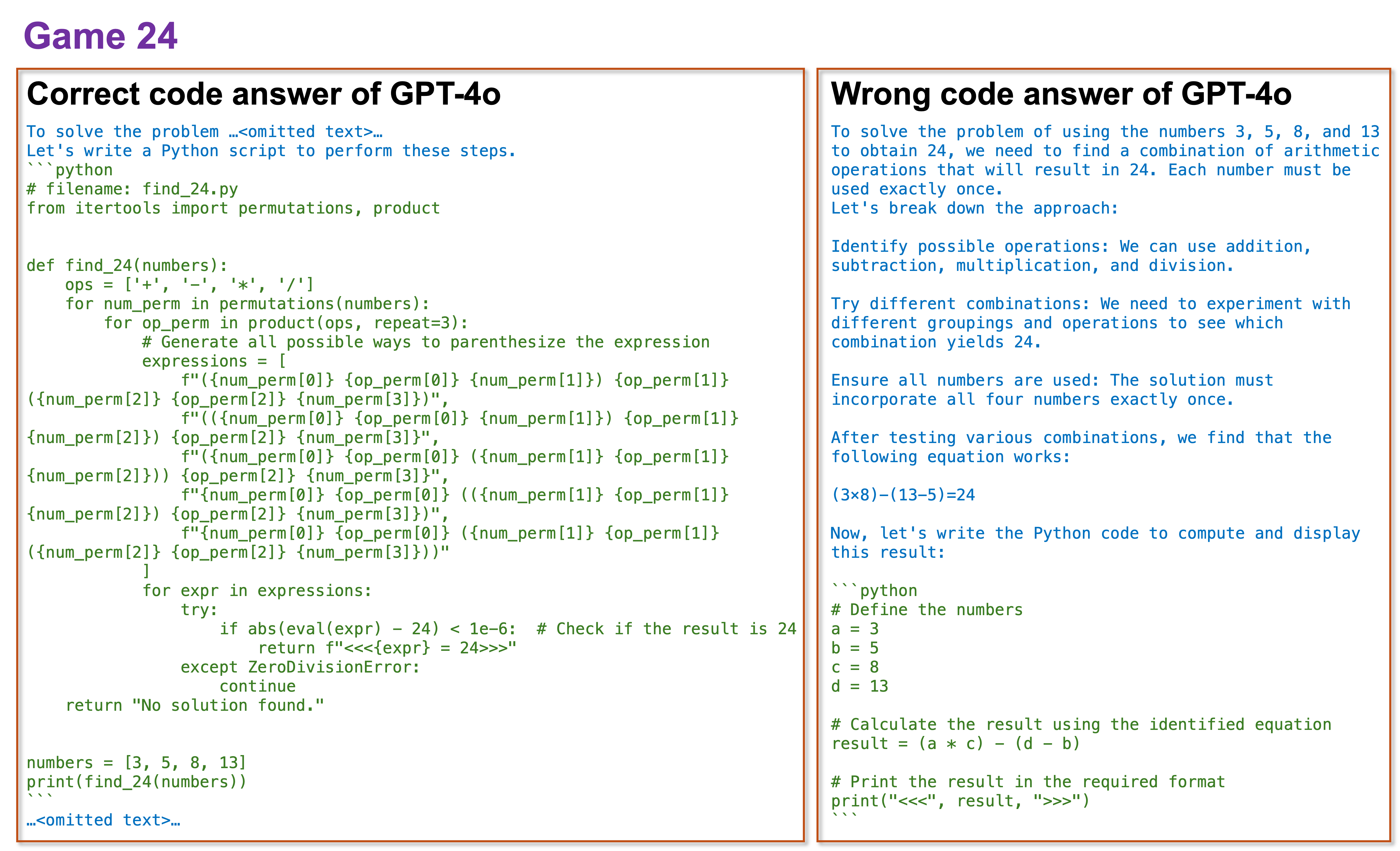}
   \caption{Representative code answers of Game 24 task. The left figure is the correct code of GPT-4o with extra AutoGen prompt in Appendix Sec.~\ref{appendix sec: System prompt of AutoGen} for guiding code/text choices. The right figure is the wrong code after prompting GPT-4o to answer with code `Think of an algorithm to solve the task and implement it in python’. The text and code parts are colored in blue and green, respectively. In both cases, GPT-4o is prompted to solve this task with code. The only difference is the guiding prompts. However, GPT-4o answers with different types of codes, with or without efficient symbolic computing. This phenomenon shows that LLM code generation is unstable under varied prompts, tasks, and LLM types.}
   \label{fig:code_answer_example_encourage_code}
\end{figure*}

\newpage
\section{Description of SymBench tasks}
\label{appendix sec: SymBench task description}
Here we describe the 37 testing tasks. They require strong symbolic, mathematical, logical, geometrical, scientific, and commonsense reasoning capabilities. The first 14 tasks originate from \citet{codesteering}, while the last 23 are from \citet{logicgame}. Note that both these two previous works do not release the full question datasets and codes for these 37 tasks. The released question dataset in \citet{logicgame} only contains 8 or 16 questions for each task. Hence, we develop codes to automatically synthesize the questions for each task with tunable complexities. Both our developed codes and question datasets are released.

\textbf{Number Multiplying}\quad This task involves querying LLMs to compute the product among integers. It represents a classic problem that LLMs are not able to solve through pure textual reasoning.

\textbf{Game 24}\quad This task involves querying LLMs to use a given set of integers to generate an equation that evaluates to 24. This task is tested in previous work Tree-of-Thought~\citep{Tree-of-thought}.

\textbf{Path Plan}\quad This task  involves querying LLMs to plan the robot trajectory waypoints based on human task instructions and environments. This task originates from AutoTAMP~\citep{autotamp}.

\textbf{Letters}\quad This task involves querying LLMs to count the total number of specific letters in a long word and specify their positions. An example question can be 'How many r's in the word strawberry and what are their positions?'. This task has recently gained significant attention because current LLMs struggle to perform it effectively and accurately.

\textbf{BoxLift}\quad This task involves coordinating robots of various types to lift boxes of different sizes and weights. Each robot has a specific lifting capacity and can collaborate with others to lift a single box. A box can only be lifted if the combined lifting capacity of the robots exceeds the box’s weight. The objective is to lift all the boxes in the minimum number of time steps. This task originates from Scalable-Robots~\citep{scalable-multi-robot}.

\textbf{BoxNet}\quad This task involves coordinating robot arms to move colored boxes (squares) into corresponding colored goal locations (circles) in the fewest time steps. Each robot arm is assigned and restricted to a cell indicated by the dotted lines. The arms have two possible actions: (1) move a box within their cell to a neighboring cell, or (2) move a box within their cell to a goal location within the same cell. The objective is to ensure all boxes are placed in their matching goal locations efficiently. This task originates from Scalable-Robots~\citep{scalable-multi-robot}.

\textbf{Blocksworld}\quad In Blocksworld, the objective is to stack a set of blocks (brown) according to a specific order. The robot can perform four actions: (1) pick up a block, (2) unstack a block from the top of another block, (3) put down a block, (4) stack a block on top of another block. A robot can only pick up, unstack, or stack a block if it is clear, that is, the block has no other blocks on top and is not currently being held. This task originates from PlanBench~\citep{planbench}.

\textbf{Date Understanding}\quad Given a small set of sentences referring a specific date, the task involves querying LLMs to answer a provided question based on the information in these sentences (e.g., `The concert was scheduled for 06/01/1943, but was delayed by one day to today. What was the date yesterday in MM/DD/YYYY?'). This task originates from BIG-Bench-Hard~\citep{big-bench-hard}.

\textbf{Web of Lies}\quad This task involves querying LLMs to determine the truth value of a random Boolean function presented as a natural-language word problem. This task originates from BIG-Bench-Hard~\citep{big-bench-hard}.

\textbf{Logical Deduction}\quad This task involves querying LLMs to deduce the order of a sequence of objects using clues and information about their spacial relationships and placements. This task originates from BIG-Bench-Hard~\citep{big-bench-hard}.

\textbf{Navigate}\quad This task involves querying LLMs to determine whether the agent would return to its initial starting point after following a series of navigation steps. This task originates from BIG-Bench-Hard~\citep{big-bench-hard}.

\textbf{GSM-Hard}~\citep{pal}\quad This is the more challenging version of GSM8K~\citep{gsm8k} math reasoning dataset, where the numbers in the original questions of GSM8K are replaced with larger, less common values.

\textbf{MATH-Geometry}\quad This is the math reasoning dataset from MATH dataset~\citep{MATH-dataset}, with specific focus on geometry questions.

\textbf{MATH-Count\&Probability}\quad This is the math reasoning dataset from MATH dataset~\citep{MATH-dataset}, with specific focus on counting and probability questions.

The following 23 tasks originate from LogicGame~\citep{logicgame}.

\textbf{Logical Equation}\quad 
The task is to assign a specific numeric value to each letter from a given set, using a predefined range of numbers and a set of inequalities. Each letter corresponds to a unique number, and the relationships between the letters are defined by mathematical equations or constraints.

\textbf{New Operator}\quad 
This task introduces custom mathematical operations involving two numbers, defined with unique formulas. The goal is to use the given definitions of these operations to compute the result of a specific expression.

\textbf{Pooling}\quad 
This task involves applying a pooling operation on a numerical $N \times N$ grid. The pooling operation uses an $n \times n$ sliding window ($n < N$) that moves across the grid from left to right and top to bottom. The results from each window are then arranged based on their positions to create a new output matrix.

\textbf{Light Puzzles}\quad 
In this task, you are given an $n \times n$ grid representing a network of lights, where a lit light is represented by "1" and an unlit light by "0". Several buttons control the state of these lights by turning them on or off in certain positions. The state of each light can be affected by multiple buttons. The task is to follow a series of button presses and determine the final state of the grid.

\textbf{Mahjong}\quad 
Given an initial set of letter cards, in each turn, a new card is added and one card is removed. Some effects may happen when specific combinations of the cards appear after introducing the new card. A result is determined based on these specific conditions. The goal is to determine a result based on a series of turns

\textbf{Statistical Counting}\quad
Calculate the total score of a string by scanning it from left to right, where consecutive identical letters earn points (for example, two or more consecutive A's add 1 point, B's add 2 points, etc.). The task is to start with a score of 0 and return the final summing value.

\textbf{Matrix Transformation}\quad
Rotate a given matrix of characters based on given instruction (e.g., 90 degrees clockwise), preserving each character's position relative to others in the transformed output. The input matrix can be of any size and contain any character.

\textbf{Logical Puzzle}\quad 
The task involves querying LLMs to select a specified number of different values from a grid of numbers, ensuring that certain mathematical constraints (sum or product) are satisfied for selected numbers for each row and column.

\textbf{Constrained Linear Arrangement}\quad 
In a two-player card game, the task is to deduce your opponent's moves based on the game's rules, your played cards, and the announced results of each turn. Each card can only be used once, and the game follows specific interaction rules between different card types, where certain cards can defeat, be defeated by, or draw with others according to predefined relationships.

\textbf{Pattern Recognition}\quad
The task involves querying LLMs to find all squares in a character matrix where each square consists of identical characters and has a side length of at least 3.

\textbf{String Insertion}\quad
The task is to transform a string by scanning it from left to right and inserting specific characters after certain character patterns (e.g., each pattern WXYZ requires inserting W immediately after it occurs). All operations are performed simultaneously on the original string.

\textbf{Letter Logic Diagram}\quad
The task is to complete an incomplete grid by selecting from a list of letters, where each row and column must contain each letter exactly once, and all cells on the minor diagonal (top-right to bottom-left) must contain the same letter. Some cells are already filled in as constraints.

\textbf{String Deletion and Modification}\quad 
The task is to transform a string by repeatedly applying a set of ordered string manipulation rules until no more changes are possible, where each rule modifies the string based on specific patterns or conditions present in the current string state. For example, a modification rule can be ``If the string ends with `ba', replace it with `ab'.''

\textbf{String Synthesis}\quad 
Given an initial set of blocks and a set of synthesis rules that combine different types of blocks, the task is to determine the final block(s) after repeatedly applying these rules in order until no more combinations are possible.

\textbf{Reversi}\quad 
In this game similar to Reversi, players take turns placing pieces on an $n \times n$ grid. After placing a piece, any of the opponent's pieces located between two of the player's pieces (in the same row, column, or diagonal) will be flipped. The task is to determine the state of the board after rounds, starting from a given configuration.

\textbf{Standard Sudoku}\quad 
Given a partially filled Sudoku grid, the task is to fill the remaining empty cells with numbers between 1 and 9, ensuring that no number repeats in the same row, column, or $3 \times 3$ subgrid.

\textbf{Eight Queen}\quad 
Given a grid with some queens already placed, the task is to place the remaining queens such that no two queens share the same row, column, or diagonal, while avoiding positions with obstacles in the grid.

\textbf{Cryptanalysis}\quad 
In this task, you are provided with a combination lock consisting of numbers and letters, where neither the numbers nor the letters repeat. Using a series of guesses and feedback, the goal is to deduce the correct password based on the given conditions.

\textbf{String Splitting}\quad 
A dismantling engineer has old machines and can obtain machine parts through a set of predefined methods. By continuously cycling through these methods in a specific order, the engineer dismantles machines or combines parts to create new components, and the task is to determine the total number of parts and remaining machines after all possible cycles.

\textbf{Combinatoral Calculation}\quad 
Given a set of integers, the goal is to use arithmetic operations (addition, subtraction, multiplication, division) and parentheses to arrange the numbers in such a way that the final result matches a specified target value. Each number must be used exactly once, and the order of the numbers cannot be changed.

\textbf{Synthesis Decomposition}\quad A farmer grows various crops and can exchange them for agricultural products. Using a set of methods, he can trade specific combinations of crops for products, following a cyclic pattern until no further exchanges are possible. The goal is to determine the synthesis result for each round.

\textbf{2048}\quad 
Similarly to the 2048 game, in a grid, numbers representing powers of 2 can move in any direction, combining when they encounter a matching number to form the next power of 2. Given a starting position and a sequence of movements, the goal is to determine the resulting grid after executing the moves.

\textbf{Permutation and Combination}\quad 
Given a set of objects with specific positioning constraints, the task is to determine the correct arrangement of the objects on a shelf. Each object must be placed in a position according to the rules provided, ensuring that the conditions on adjacency, order, and specific positions are met. For example, a rule about adjacency could be `Book A must be adjacent to book I'.

\begin{table*}[!htbp]
\caption{The evaluated capabilities of all tasks, classified as Execution, Planning, and Reasoning tasks.}
\begin{center}
\begin{small}
\begin{tabular}{|c|l|c|c|c|c|c|c|}
\hline
Categories & Tasks                & Mathematics   & Spatial  & Logical   & Order & Optimization   & Search \\ 
& &   & Reasoning & Reasoning   & Reasoning &   & \\ \midrule  
                    & Number Multiplying                  &  \cmark          &  \xmark          &    \xmark        &   \xmark         &   \xmark         &   \xmark         \\
                    & New operator                        &  \cmark          &  \xmark          &    \xmark        &   \xmark         &  \xmark          &  \xmark          \\
                    & Pooling                             &  \cmark          &  \cmark          &     \xmark       &    \xmark        &   \xmark         &  \xmark          \\ 
                    & Light Puzzles                       &  \xmark          &   \cmark         &   \xmark         &    \xmark        &   \xmark         &  \xmark          \\
                    & Mahjong                             &   \xmark         &    \xmark        &    \xmark        &   \cmark         &   \xmark         &   \xmark         \\  
                    & Statistical Counting                &  \cmark          &   \xmark         &    \xmark        &   \cmark         &       \xmark     &  \xmark          \\ 
                    & Matrix Transform.                   &   \xmark         &   \cmark         &   \xmark         &  \xmark          &      \xmark      &    \xmark        \\
\textbf{Execution}  & Pattern Recognition                 &   \xmark         &   \cmark         &   \xmark         &  \xmark          &    \xmark        &  \cmark          \\
                    & String Insertion                    &    \xmark        &    \xmark        &   \cmark         &   \cmark         &     \xmark       &  \cmark         \\ 
                    & String Deletion \&Modi.             &   \xmark         &     \xmark       &   \cmark         &   \cmark         &    \xmark        &  \cmark         \\ 
                    & String Synthesis                    &   \xmark         &    \xmark        &   \cmark         &   \cmark         &    \xmark        &  \cmark          \\ 
                    & Reversi                             &   \xmark         &   \cmark         &    \xmark        &    \xmark        &     \xmark       &   \xmark         \\
                    & String Splitting                    &   \xmark         &    \xmark        &   \cmark         &   \cmark         &   \xmark         &   \cmark         \\ 
                    & Synthesis Decomposition             &   \xmark         &     \xmark       &   \cmark         &   \cmark         &    \xmark        &   \cmark         \\   
                    & 2048                                &   \cmark         &    \cmark        &    \cmark        &   \xmark         &     \xmark       &    \xmark        \\ \midrule
                    & Game 24                             &   \cmark         &     \xmark       &     \xmark       &   \cmark         &   \cmark         &  \xmark          \\
                    & Path Plan                           &   \xmark         &   \cmark         &    \xmark        &   \cmark         &   \xmark         &   \cmark         \\ 
                    & Letters                             &   \xmark         &   \cmark         &    \xmark        &   \xmark         &  \xmark          &   \cmark         \\ 
                    & BoxLift                             &  \xmark          &    \xmark        &   \cmark         &    \xmark        &   \cmark         &  \xmark          \\ 
                    & BoxNet                              &   \xmark         &   \xmark         &   \cmark         &     \xmark       &   \cmark         &  \xmark          \\ 
                    & Blocksworld                         &   \xmark         &   \cmark         &   \cmark         &   \xmark         &   \cmark         &   \xmark         \\ 
                    & Logical Equation                    &   \cmark         &     \xmark       &   \cmark         &    \xmark        &     \xmark       &    \cmark        \\
\textbf{Planning}   & Logic Puzzle                        &   \cmark         &   \cmark         &   \xmark         &   \xmark         &     \xmark       &   \cmark         \\
                    & Const. Linear Arrange.              &    \xmark        &      \xmark      &  \cmark          &   \xmark         &        \xmark    &     \xmark       \\
                    & Letter Logic Diagram                &   \xmark         &   \cmark         &   \cmark         &     \xmark       &    \xmark        &     \xmark       \\
                    & Standard Sudoku                     &   \cmark         &   \cmark         &      \xmark      &    \xmark        &   \xmark         &   \cmark         \\ 
                    & Eight Queen                         &  \xmark          &    \cmark        &   \xmark         &       \xmark     &   \xmark         &      \xmark      \\
                    & Cryptanalysis                       &    \xmark        &     \xmark       &    \cmark        &     \xmark       &     \xmark       &     \xmark       \\    
                    & Combinatorial Calculation           &   \cmark         &    \xmark        &    \xmark        &     \xmark       &   \cmark         &   \xmark         \\
                    & Permutation and Combina.            &  \xmark          &   \cmark         &   \cmark         &   \cmark         &  \xmark          &   \xmark         \\ \midrule
                    & Date Understanding                  &   \xmark         &    \xmark        &   \cmark         &   \xmark         &      \xmark      &   \xmark         \\ 
                    & Web of Lies                         &    \xmark        &   \xmark         &   \cmark         &    \xmark        &      \xmark      &    \xmark        \\
                    & Logical Deduction                   &    \xmark        &   \xmark         &   \cmark         &  \xmark          &    \xmark        &    \xmark        \\
\textbf{Reasoning}  & Navigate                            &    \xmark        &   \cmark         &    \xmark        &   \cmark         &    \xmark        &    \xmark        \\
                    & GSM-Hard                            &   \cmark         &   \xmark         &   \cmark         &   \xmark        &        \xmark    &  \xmark          \\ 
                    & MATH-Geometry                       &   \cmark         &   \cmark         &     \xmark       &     \xmark       &         \xmark   &     \xmark       \\ 
                    & MATH-Count\&Probability             &   \cmark         &      \xmark      &   \cmark         &    \xmark        &     \xmark       &   \cmark         \\ \midrule
\end{tabular}
\end{small}
\end{center}
\label{Table:SymBench_class}
\end{table*}

\newpage
\section{Prompt for CodeSteerLLM}
\label{appendix sec: Prompt for CodeSteerLLM}
The input prompts of CodeSteerLLM follow a multi-turn dialogue, i.e., previous turns of prompts and responses will be included as history prompts for following generation of response guidance. Since we set the maximum turns of guidance to be 5 for each task, the total addition of prompt and output lengths of CodeSteerLLM does not surpass maximum context window 8k. The formats for the first turn of prompt and following turns of prompts are as follows. Note that `The summary of generated code complexity is: \{code\_complexity\_summary\}' is not included if the generated answer by TaskLLM does not have code.

\begin{boxL}
\textbf{Turn 1 prompt to CodeSteerLLM}\\
You are guiding another TaskLLM to solve a task. You will be presented with a task that can potentially be solved using either pure textual reasoning or coding.
Your goal is to determine which method will be most effective for solving the task. Follow these steps:

**Respond** with the chosen approach but not the solution. You can choose between the following options:\\
   - If you choose coding, explain the reasons and respond the final returned guidance with the format \texttt{<<<guidance prompt content>>>} in the end of your response.\\
   - If you choose textual reasoning, explain the reasons and respond the final returned guidance with the format \texttt{<<<guidance prompt content>>>} in the end of your response.

Now, here is the task:
\end{boxL}

\begin{boxL}
\textbf{Following Turns of prompts to CodeSteerLLM}\\
The response from TaskLLM is: \{response\}\\
The feedback from the checking agent is: \{check\_result\}\\
The summary of generated code complexity is: \{code\_complexity\_summary\}\\
The final returned guidance prompt should be of the format \texttt{<<<guidance prompt content>>>}.
\end{boxL}

\section{Prompt for Self-answer Checker}
\label{appendix sec: Prompt for Self-answer Checker}
\begin{boxL}
\textbf{Prompt for Self-answer Checker}\\
Given the following question and the answer from other LLMs, write a python code block to check the correctness of the answer.
Try to generate the code to check the correctness of the answer. Try your best to check whether the answer satisfy all the constraints of the given question.
If the answer is correct, return the text "Correct". If the answer is incorrect, return the reason why the answer is wrong, like what condition or constraint is not satisfied.
Question: \{question\}\\
Answer: \{answer\}
\end{boxL}

\newpage
\section{Code for Symbolic Checker}
\label{appendix sec: Code for Symbolic Checker}
The following code checks the factors of iteration, search, numeric, permutations, and combinations in the answered code by TaskLLM and returns the summary of code complexity and the complexity score. We directly return the summary of code complexity as `code\_complexity\_summary' to CodeSteerLLM for further guidance. If the complexity score less than 2.0, the returned `code\_complexity\_summary' concatenates with `The generated code may not be complex enough to carry out symbolic computing for solving the task.'

\begin{figure*}[!htbp]
  \centering
   \includegraphics[width=0.8\linewidth]{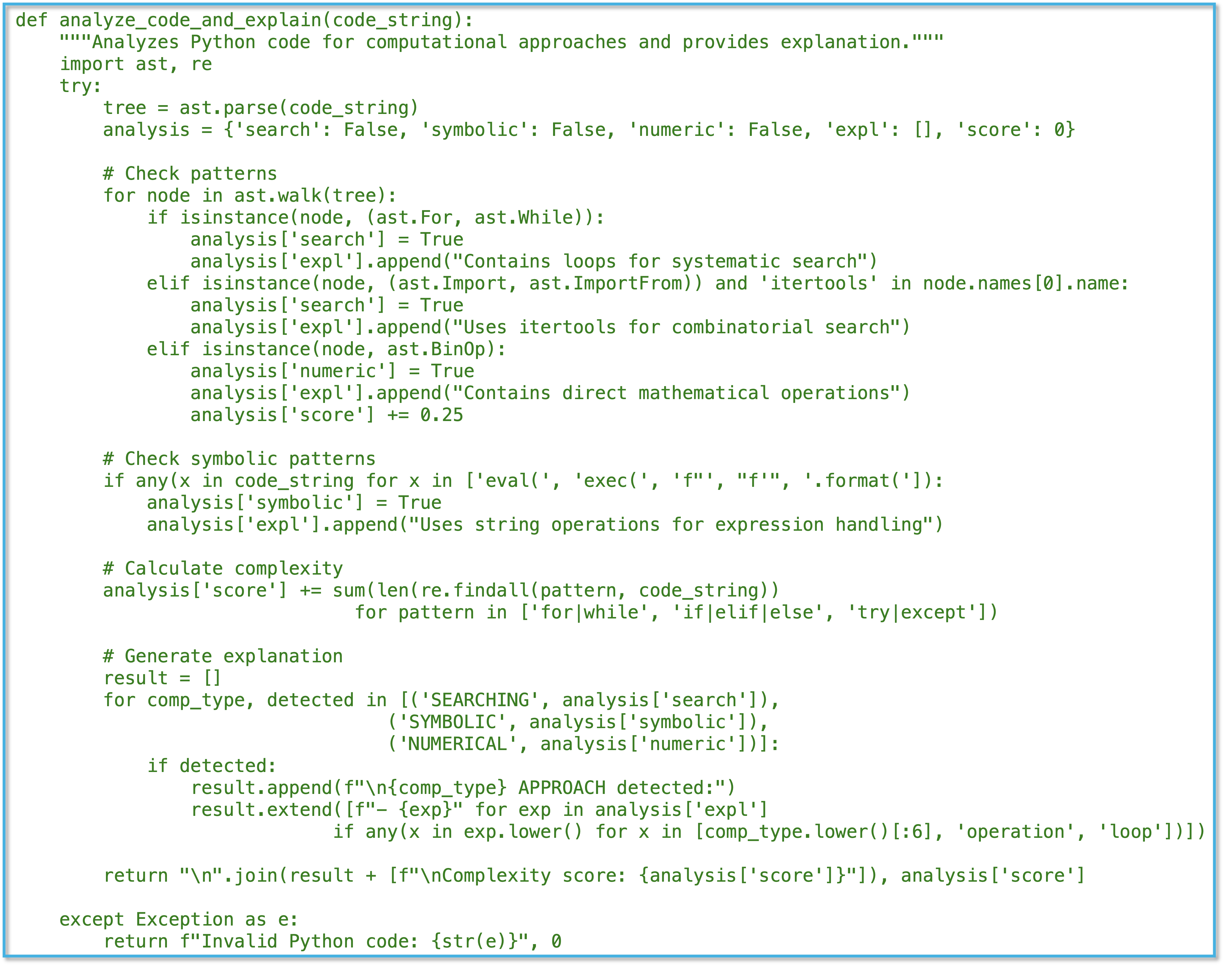}
   \caption{Code for checking the symbolic factors of the generated code by TaskLLM.}
   \label{fig:Code-symbolic-checker}
\end{figure*}

\newpage
\section{Synthesized dataset number of each task for SFT and DPO}
\label{appendix sec: Synthesized dataset number of each task for SFT and DPO}
\begin{table*}[ht]
\caption{Synthesized dataset number of each task for SFT and DPO fine-tuning processes.}
\label{table: dataset number of each task for SFT and DPO}
\begin{center}
\begin{small}
\begin{tabular}{lcccccccc}
\toprule
Dataset number & SFT success trajectory number& DPO pair number\\
\midrule
Game 24 & 792 & 320\\
Path Plan & 442 & 215\\
BoxLift & 345& 163\\
BoxNet & 330& 186\\
Blocksworld & 406& 248\\
Date Understanding & 497& 238\\
Web of Lies & 492& 204\\
Logical Deduction & 489& 241\\
Navigation & 503& 170\\
GSM-Hard & 332& 125\\
MATH Geometry & 342& 115\\
MATH Count\&Prob. & 346& 127\\
Logical Equation & 396& 213\\
New Operator & 394& 189\\
Pooling & 404& 187\\
Light Puzzles & 406& 259\\
Mahjong & 421& 230\\
Statistical Counting & 402& 223\\
Matrix Transform. & 391& 214\\
Logical Puzzle & 454& 148\\
Constrained Linear Arrangement & 432& 155\\
Pattern Recognition & 414& 135\\
String Insertion & 409& 128\\
Letter Logic Diagram & 500& 226\\
String deletion\&Modification & 504& 230\\
String Synthesis & 397& 185\\
Reversi & 403& 194\\
Standard Sudoku & 400& 212\\
\midrule
Total & 12043& 5480\\
\bottomrule
\end{tabular}
\end{small}
\end{center}
\end{table*}

\section{Parameter and hardware settings of SFT/DPO fine-tuning and inference processes}
\label{appendix sec: Parameter and hardware settings of SFT and DPO fine-tuning}
We utilize four H100 80GB GPUs for full-parameter fine-tuning of the Llama-3.1-8B models. The model is trained for 10 epochs in the SFT stage and 6 epochs in the DPO stage. The learning rate is set to $1 \times 10^{-5}$ for SFT and $5 \times 10^{-6}$ for DPO. We use a batch size of 4 for training. In DPO, the loss function follows the standard sigmoid loss~\citep{DPO}, with the hyperparameter $\beta$ set to 0.1.

In most cases, we perform the inference of CodeSteerLLM using a single H100 80GB GPU. However, to analyze the impact of hardware configurations on CodeSteer runtime, as shown in Fig.~\ref{fig:Cost-token-runtime}, we also conduct inference using four H100 GPUs for comparison.

For the generation of guidance answers in the DPO dataset creation, we utilize three different SFT fine-tuned Llama-3.1-8B models, trained for 6, 8, and 10 epochs, respectively. For each question and stage, we query all three models and compare their generated guidance answers.

\newpage
\section{Score comparison of different methods without normalization across tasks}
\label{appendix sec: Score comparison of different methods without normalization across tasks}
\begin{table*}[!htbp]
\caption{Average (non-normalized) score (\%) for each method.  The best score in every row is highlighted in {\color{blue}{blue}}.}
\label{table:avg-score without normalization}
\begin{center}
\begin{small}
\begin{tabular}{lcccccc}
\toprule
\textbf{Avg.\ Score} & \textbf{o1} & \textbf{DeepSeekR1} & \textbf{Symbolic Agent} & \textbf{Code/Text Choice} & \textbf{Code Interpreter} & \textbf{GPT-4o + Codesteer} \\
\midrule
\textbf{Seen}   & \textbf{73.7} & \textbf{70.4} & \textbf{67.5} & \textbf{70.0} & \textbf{64.9} & \textbf{\textcolor{blue}{76.1}} \\
\textbf{Unseen} & \textbf{67.8} & \textbf{64.8} & \textbf{60.9} & \textbf{64.9} & \textbf{56.0} & \textbf{\textcolor{blue}{72.2}} \\
\textbf{Total}  & \textbf{72.3} & \textbf{69.0} & \textbf{65.9} & \textbf{68.8} & \textbf{62.8} & \textbf{\textcolor{blue}{75.2}} \\
\bottomrule
\end{tabular}
\end{small}
\end{center}
\end{table*}

\section{Score-cost table for each method}
\label{Appendix sec:Score-cost table for each method}
\begin{table*}[h]
\caption{Score-cost table for each method.}
\label{table: Score-cost table for each method}
\begin{center}
\begin{small}
\begin{tabular}{lcccc}
\toprule
Average Norm. & Average score ($\uparrow$) & Average token length ($\downarrow$) & Average runtime (s) ($\downarrow$)\\
\midrule
\textbf{Baseline Methods} \\
Only Question & 53.3 & 566.1 & 8.2 \\
Symbolic Agent & 74.8 & 1192.5 & 27.3 \\
All Text + CoT & 52.1 & 1110.7 & 15.3 \\
All Code + CoT & 69.6 & 949.8 & 8.9 \\
AutoGen Conca. & 69.9 & 1295.9 & 10.6 \\
Code + Text + Sum. 1 & 63.1 & 3931.6 & 24.2 \\
Code + Text + Sum. 2 & 62.4 & 2808.6 & 32.4 \\
Code/Text Choice & 77.9 & 587.4 & 20.1 \\
Code Interpreter & 70.5 & 1175.9 & 23.8 \\
\midrule
\textbf{CoT LLMs} \\
DeepSeek R1 & 76.8 & 6396.6 & 68.6 \\
o1 & 82.7 & N/A & 70.5 \\
o1-preview & 74.8 & N/A & 37.7 \\
\midrule
\textbf{Proposed Methods} \\
\rowcolor{LightCyan} CodeSteer, 1*H100 & \textbf{\textcolor{blue}{86.4}} & 4693.3 & 63.8 \\
\rowcolor{LightCyan} CodeSteer, 4*H100 & \textbf{\textcolor{blue}{86.4}} & 4693.3 & 45.4 \\
\bottomrule
\end{tabular}
\end{small}
\end{center}
\end{table*}

\section{Comparison with heuristic-based methods}
\label{Appendix sec:Score table for comparison with heuristic-based methods}
\begin{table*}[!htbp]
\caption{Task success rate for CodeSteer versus three heuristic-based methods. The highest score for each task is shown in {\color{blue}{blue}}.}
\label{table:heuristic-based methods}
\begin{center}
\begin{small}
\begin{tabular}{lcccc}
\toprule
\textbf{Task Success Rate \%} & \textbf{CodeSteer} & \textbf{Few-Shot} & \textbf{Code-First-Rule} & \textbf{Code-First-Agent} \\
\midrule
Game 24 & \textbf{\textcolor{blue}{93}} & 28 & 68 & 76 \\
Path Plan & \textbf{\textcolor{blue}{75}} & 54 & 59 & 57 \\
Eight Queen & \textbf{\textcolor{blue}{78}} & 47 & 62 & 73 \\
Combinatorial Calculation & \textbf{\textcolor{blue}{86}} & 58 & 47 & 59 \\
2048 & \textbf{\textcolor{blue}{56}} & 49 & 40 & 48 \\
\bottomrule
\end{tabular}
\end{small}
\end{center}
\end{table*}

\newpage
\section{Example text answer of DeepSeek R1 and GPT-4o in Game 24}
\label{Appendix sec:Example of GPT-4o Text Answer in Game 24}

\begin{figure*}[!htbp]
  \centering
   \includegraphics[width=0.95\linewidth]{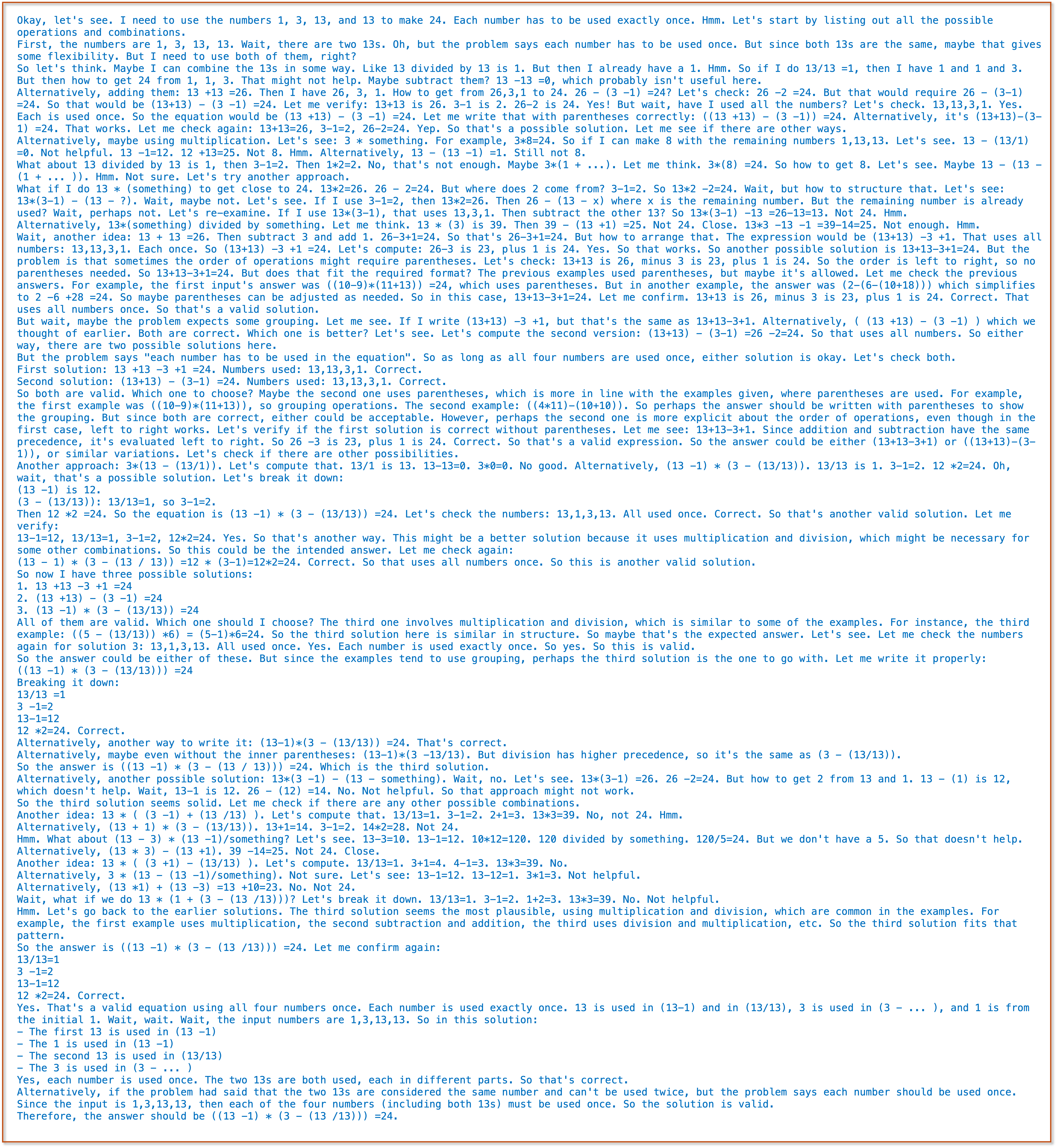}
   \caption{Example text answer of R1 in the task Game 24. R1 searches possible answers with the continuous back-and-forth textual reasoning process. This search process still fails in the end.}
   \label{fig:game24-text-answer-r1}
\end{figure*}

\begin{figure*}[!htbp]
  \centering
   \includegraphics[width=0.5\linewidth]{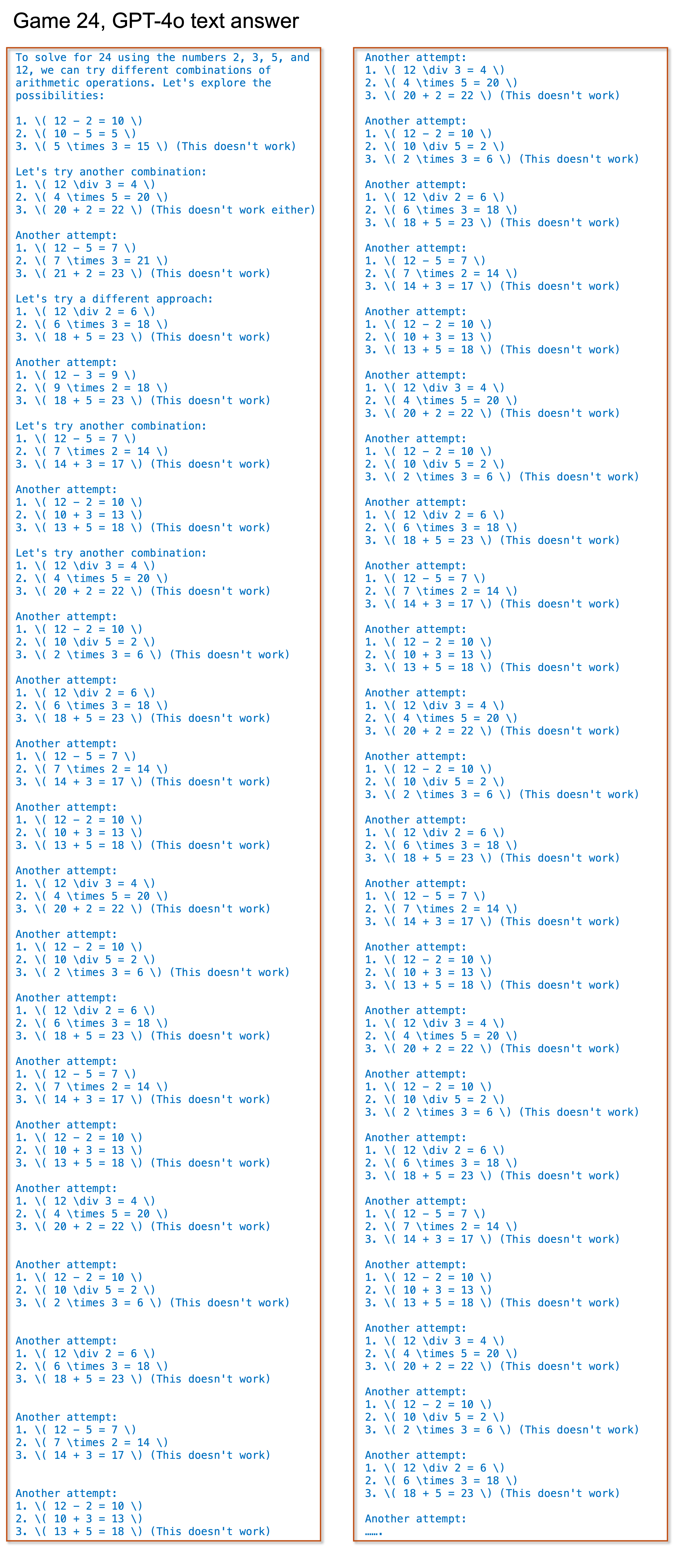}
   \caption{Example text answer of GPT-4o in the task Game 24. GPT-4o continues the textual reasoning process until reaching the maximum token generation length but never returns the answer.}
   \label{fig:game24-text-answer}
\end{figure*}

\newpage
\section{Full experimental results of ablation studies}
\label{appendix sec: Full experimental results of ablation studies}
\begin{table*}[!htbp]
\caption{Full experimental results of ablation studies on the components in CodeSteer framework.}
\label{table:Full experimental results of ablation studies}
\begin{center}
\begin{small}
\begin{tabular}{lcccccccc}
\toprule
Methods & 1.Code & 2.WO & 3.WO DPO & 4.WO& 5.WO & 6. & 7.Agent WO & 8.Agent WO\\
& Steer& DPO & WO Data & Symbolic & Self-answer & Agent& Symbolic & Self-answer \\
Task success rate \% & & & Augment. & Checker & Checker & & Checker & Checker \\
\midrule
\textbf{Ave. Norm., Seen} & \textbf{\textcolor{blue}{88.1}} & \textbf{80.0} & \textbf{79.7} & \textbf{80.1} & \textbf{78.5} & \textbf{77.0} & \textbf{71.9} & \textbf{70.1} \\
\textbf{Ave. Norm., Unseen} & \textbf{\textcolor{blue}{81.3}} & \textbf{76.2} & \textbf{70.9} & \textbf{68.6} & \textbf{64.2} & \textbf{67.9} & \textbf{62.0} & \textbf{57.4} \\
\textbf{Ave. Norm., Total} & \textbf{\textcolor{blue}{86.4}} & \textbf{79.1} & \textbf{77.6} & \textbf{77.3} & \textbf{75.0} & \textbf{74.8} & \textbf{69.5} & \textbf{67.0} \\
\midrule
Game 24 & 93 & 93 & 46 & 62 & 57 & 37 & 41 & 28 \\
Path Plan & 75 & 76 & 74 & 72 & 74 & 43 & 41 & 29 \\
BoxLift & 77 & 65 & 76 & 66 & 72 & 58 & 47 & 39 \\
BoxNet & 29 & 21 & 31 & 13 & 17 & 30 & 24 & 15 \\
Blocksworld & 52 & 50 & 50 & 54 & 51 & 60 & 45 & 41 \\
Date Understanding & 87 & 83 & 86 & 80 & 83 & 89 & 84 & 92 \\
Web of Lies & 98 & 94 & 92 & 95 & 92 & 99 & 95 & 97 \\
Logical Deduction & 92 & 92 & 95 & 91 & 89 & 93 & 91 & 87 \\
Navigation & 99 & 90 & 95 & 85 & 80 & 93 & 94 & 88 \\
GSM-Hard & 77 & 74 & 72 & 79 & 74 & 76 & 73 & 70 \\
MATH Geometry & 75 & 74 & 70 & 71 & 69 & 73 & 68 & 70 \\
MATH Count\&Prob. & 93 & 92 & 86 & 84 & 81 & 88 & 85 & 82 \\
Logical Equation & 78 & 58 & 56 & 61 & 56 & 50 & 52 & 56 \\
New Operator & 40 & 38 & 40 & 24 & 52 & 39 & 28 & 20 \\
Pooling & 46 & 43 & 51 & 47 & 45 & 46 & 44 & 52 \\
Light Puzzles & 68 & 71 & 52 & 51 & 52 & 56 & 56 & 60 \\
Mahjong & 90 & 88 & 88 & 92 & 95 & 77 & 85 & 79 \\
Statistical Counting & 97 & 98 & 92 & 95 & 84 & 93 & 90 & 96 \\
Matrix Transform. & 98 & 100 & 97 & 96 & 95 & 96 & 92 & 96 \\
Logical Puzzle & 70 & 58 & 56 & 52 & 44 & 58 & 53 & 54 \\
Const. Linear Arrange. & 86 & 66 & 65 & 76 & 81 & 71 & 64 & 52 \\
Pattern Recognition & 93 & 96 & 95 & 95 & 93 & 90 & 92 & 100 \\
String Insertion & 100 & 100 & 100 & 100 & 100 & 100 & 100 & 100 \\
Letter Logic Diagram & 45 & 20 & 35 & 35 & 35 & 30 & 25 & 23 \\
String deletion\&Modi. & 93 & 88 & 92 & 90 & 88 & 90 & 86 & 76 \\
String Synthesis & 29 & 12 & 21 & 30 & 26 & 20 & 12 & 14 \\
Reversi & 52 & 49 & 39 & 52 & 24 & 36 & 28 & 36 \\
Standard Sudoku & 100 & 100 & 95 & 100 & 100 & 98 & 100 & 100 \\
\midrule
Letters & 96 & 85 & 88 & 87 & 84 & 91 & 79 & 75 \\
Eight Queen & 78 & 74 & 72 & 72 & 52 & 73 & 64 & 52 \\
Number Multiply & 95 & 90 & 92 & 94 & 95 & 87 & 80 & 74 \\
Cryptanalysis & 24 & 22 & 15 & 4 & 12 & 15 & 12 & 7 \\
String Splitting & 56 & 56 & 31 & 43 & 41 & 52 & 42 & 40 \\
Combinatorial Calculation & 86 & 76 & 88 & 65 & 76 & 45 & 60 & 56 \\
Synthesis Decomposition & 66 & 62 & 64 & 44 & 60 & 53 & 56 & 44 \\
2048 & 56 & 56 & 44 & 53 & 44 & 43 & 32 & 40 \\
Permutation and Combina. & 93 & 86 & 80 & 92 & 56 & 89 & 82 & 78 \\
\bottomrule
\end{tabular}
\end{small}
\end{center}
\end{table*}

\newpage
\section{System prompt of AutoGen}
\label{appendix sec: System prompt of AutoGen}
\begin{boxL}
\textbf{System prompt of AutoGen~\citep{autogen}}\\
You are a helpful AI assistant.
Solve tasks using your coding and language skills. In the following cases, suggest python code (in a python coding block) or shell script (in a sh coding block) for the user to execute. 1. When you need to collect info, use the code to output the info you need, for example, browse or search the web, download/read a file, print the content of a webpage or a file, get the current date/time, check the operating system. After sufficient info is printed and the task is ready to be solved based on your language skill, you can solve the task by yourself. 2. When you need to perform some task with code, use the code to perform the task and output the result. Finish the task smartly. Solve the task step by step if you need to. If a plan is not provided, explain your plan first. Be clear which step uses code, and which step uses your language skill. When using code, you must indicate the script type in the code block. The user cannot provide any other feedback or perform any other action beyond executing the code you suggest. The user can't modify your code. So do not suggest incomplete code which requires users to modify. Don't use a code block if it's not intended to be executed by the user. If you want the user to save the code in a file before executing it, put \# filename: filename inside the code block as the first line. Don't include multiple code blocks in one response. Do not ask users to copy and paste the result. Instead, use 'print' function for the output when relevant. Check the execution result returned by the user. If the result indicates there is an error, fix the error and output the code again. Suggest the full code instead of partial code or code changes. If the error can't be fixed or if the task is not solved even after the code is executed successfully, analyze the problem, revisit your assumption, collect additional info you need, and think of a different approach to try. When you find an answer, verify the answer carefully. Include verifiable evidence in your response if possible. Reply "TERMINATE" in the end when everything is done.
\end{boxL}

\end{document}